\documentclass[preprint,11pt]{elsarticle}

\usepackage{tikz}
\usetikzlibrary{arrows.meta, positioning, calc, shadows, fit, backgrounds, matrix}

\usepackage{graphicx}%
\usepackage{multirow}%
\usepackage{amsmath,amssymb,amsfonts}%
\usepackage{amsthm}%
\usepackage{mathrsfs}%
\usepackage[title]{appendix}%
\usepackage{pifont}  
\usepackage{graphicx} 
\usepackage{textcomp}%
\usepackage{manyfoot}%
\usepackage{booktabs}%
\usepackage{algorithm}%
\usepackage{algorithmicx}%
\usepackage{algpseudocode}%
\usepackage{listings}%
\usepackage{acro}
\usepackage{enumitem}
\usepackage{xspace} 
\usepackage{xfrac} 
\usepackage{array}

\usepackage{mathtools}
\usepackage{placeins}
\usepackage{todonotes}

\definecolor{tobiasgreen}{rgb}{0.55,0.71,0.0}
\definecolor{sebastianpurple}{rgb}{0.58, 0.00, 0.83}		
\definecolor{peterblue}{rgb}{0.0, 1.0, 0.5}      
\definecolor{michaelorange}{rgb}{1.0, 0.49, 0.0}  
\definecolor{johannesred}{rgb}{1.0, 0.1, 0.24}     
\definecolor{allblue}{rgb}{0.4,0.4,1.0}

\newcommand{\ba}{\begin{array}{c}} 
\newcommand{\ea}{\end{array}}

\newcommand{\be}{\begin{equation}}
\newcommand{\ee}{\end{equation}}

\newcommand{\bea}{\begin{eqnarray}}
\newcommand{\eea}{\end{eqnarray}}

\newcommand{\bi}{\begin{itemize} \itemsep 2pt}
\newcommand{\ei}{\end{itemize}}

\newcommand{\bn}{\begin{enumerate} \itemsep 2pt}
\newcommand{\en}{\end{enumerate}}

\newcommand{\fig}[1]{Fig.~\ref{#1}}
\newcommand{\mytab}[1]{Table~\ref{#1}}

\newcommand{\refSection}[1]{Section~\ref{#1}}

\DeclareFixedFont{\ttb}{T1}{txtt}{bx}{n}{8} 
\DeclareFixedFont{\ttm}{T1}{txtt}{m}{n}{8}  

\definecolor{deepblue}{rgb}{0,0,0.5}
\definecolor{deepred}{rgb}{0.6,0,0}
\definecolor{deepgreen}{rgb}{0,0.5,0}
\definecolor{commentgreen}{RGB}{78,154,6}
\definecolor{backcolour}{rgb}{0.97,0.97,0.97}

\usepackage{tabularx}
\newcolumntype{L}[1]{>{\raggedright\let\newline\\\arraybackslash\hspace{0pt}}m{#1}}
\newcolumntype{C}[1]{>{\centering\let\newline\\\arraybackslash\hspace{0pt}}m{#1}}
\newcolumntype{R}[1]{>{\raggedleft\let\newline\\\arraybackslash\hspace{0pt}}m{#1}}

\usepackage{subcaption}

\newcommand\pythonstyle{\lstset{
language=Python,
basicstyle=\ttm,
otherkeywords={self},             
keywordstyle=\ttb\color{deepblue},
emph={AddNode, AddObject, AddMarker, AddLoad, AddSensor, SystemContainer, 
AddSystem, Assemble, SimulationSettings,
StartRenderer, StopRenderer, GetNodeOutput, function, 
TimeIntegrationSolve,__init__,
CreateGround,CreateMassPoint,CreateSpringDamper,CreateSphericalJoint
},          
emphstyle=\ttb\color{deepred},    
stringstyle=\color{deepgreen},
commentstyle=\color{commentgreen},
frame = single,
rulecolor=\color{black}, 		
showstringspaces=false,           %
numbers=left,										 
breaklines=true,									 
numberstyle=\ttm,    						
backgroundcolor=\color{backcolour}
}}

\lstnewenvironment{python}[1][]
{
\pythonstyle
\lstset{#1}
}
{}

\newcommand\pythoninline[1]{{\pythonstyle\lstinline!#1!}}

\usepackage{tikz}

\usepackage{cancel}

\theoremstyle{thmstyleone}%
\theoremstyle{thmstyletwo}%

\theoremstyle{thmstylethree}%

\definecolor{acroColor}{rgb}{0.,0.,0.}
\usepackage{hyperref} 
\hypersetup{
						colorlinks=true, 
						breaklinks,
            linkcolor=acroColor, 
            citecolor=acroColor
						}

\usepackage{acro} 
\usepackage{multicol}
\usepackage{scrextend} 
\NewAcroTemplate[list]{twocolumn}{%
  \begin{multicols}{2}[\acroheading]
    \acropreamble
    \begin{labeling}{\hspace{1.2cm}}
      \setlength{\itemsep}{1pt}
      \acronymsmapF
        {\item[{\bf\acrowrite{short}}]\acrowrite{list}}
        {\item\AcroRerun}
    \end{labeling}
  \end{multicols}
}
\acsetup{
  list/template = twocolumn,
  make-links
}
\newcommand{\myacrodef}[2]{\DeclareAcronym{#1}{short=#1, long=#2} } 

\myacrodef{AI}{Artificial Intelligence} 
\myacrodef{CoT}{Chain of Thought}
\myacrodef{DoE}{Design of Experiments}
\myacrodef{DOFs}{Degrees of Freedom}
\myacrodef{ICL}{In-Context-Learning}
\myacrodef{LFUI}{Leopold-Franzens-University Innsbruck}
\myacrodef{LLM}{Large Language Model}
\myacrodef{SLM}{Small Language Model}

\myacrodef{LoRA}{Low-Rank Adaptation}
\myacrodef{MBD}{Multibody Dynamics}
\myacrodef{fMBS}{flexible Multibody System}
\myacrodef{NLP}{Natural Language Processing}
\myacrodef{FEM}{Finite Element Method}
\myacrodef{OCR}{Optical Character Recognition}
\myacrodef{PEFT}{Parameter-Efficient Fine-Tuning}
\myacrodef{PI}{Principal Investigator}
\myacrodef{QA}{Questions and Answers}
\myacrodef{RAG}{Retrieval Augmented Generation}
\myacrodef{RL}{Reinforcement Learning}
\myacrodef{UIBK}{University of Innsbruck}
\myacrodef{WP}{Work Package}
\myacrodef{SOTA}{State-Of-The-Art}
\myacrodef{BERT}{Bidirectional Encoder Representations from Transformers}
\myacrodef{LAMMPS}{Large-scale Atomic/Molecular Massively Parallel Simulator}

\journal{TBA}

\usepackage{lineno} 
\usepackage{colortbl}
\usepackage{booktabs}
\usepackage{siunitx}
\usepackage{etoolbox}
\makeatletter
\patchcmd{\pprintMaketitle}{\vskip36pt}{\vskip20pt}{}{}
\makeatother

\begin{document}
\definecolor{primary}{HTML}{1E293B}   	
\definecolor{llmbeam}{HTML}{2563EB}   	
\definecolor{gtbeam}{HTML}{059669}    	
\definecolor{bglight}{HTML}{F8FAFC}   	
\definecolor{black}{HTML}{000000}   	

\definecolor{primary}{RGB}{50,80,140}
\definecolor{bglight}{RGB}{240,244,252}
\definecolor{mbsblue}{RGB}{90,120,190}
\definecolor{partgreen}{RGB}{60,150,90}
\definecolor{altred}{RGB}{180,70,70}
\definecolor{fixedgreen}{RGB}{40,130,70}

\begin{frontmatter}

\title{Large Language Models and their Awareness of Mechanics and Spatial Geometry}

\author[inn]{J.\ Gerstmayr\footnote{Corresponding author E-mail: \href{mailto:johannes.gerstmayr@uibk.ac.at}{johannes.gerstmayr@uibk.ac.at}}}
\author[inn]{S.\ Weyrer}
\author[inn]{T.\ Möltner}
\author[aug]{P.\ Manzl}
\author[inn]{M.\ Pieber}

\affiliation[inn]{organization={Department of Mechatronics, University of Innsbruck},
            addressline={Technikerstraße 13},
            city={Innsbruck},
            postcode={6020},
            country={Austria}}

\affiliation[aug]{organization={Institute of Materials Resource Management (MRM), University of Augsburg},
            addressline={Am Technologiezentrum 8},
            city={Augsburg},
            postcode={86159},
            country={Germany}}
            
\begin{abstract}
Large Language Models (LLMs) perform well on established code-generation and mathematical-reasoning benchmarks, but their capabilities in mechanics and spatial geometry, here denoted as mechanical engineering awareness, has not been quantified systematically. We present MecEng, a fully automated benchmark that evaluates LLMs on the creation of multibody simulation models from parameterized textual descriptions. The benchmark comprises 84 generic tasks on three difficulty levels, ranging from rigid-body systems with joints and contact to flexible multibody systems that require exact 3D geometry generation, tetrahedral finite-element meshing, and Hurty–Craig–Bampton model order reduction of machine parts. A dedicated pipeline with LLMs generates simulation-ready geometry from text using Netgen, and builds multibody system models for the code Exudyn, which are then verified against expert ground truth on several levels: system-graph isomorphism including graph node annotations, numerical solutions, and part-specific measures such as mass, geometry, and eigenfrequencies. In total, 32 open-weight and two proprietary LLMs are evaluated. On rigid-body tasks, the best open-weight model obtains an overall success rate of 86.0\%, compared to 91.4\% for the strongest proprietary model, while flexible multibody tasks remain considerably harder. Additional studies quantify the influence of sampling temperature, reasoning, prompt design, model size, and LLM-release date. The results indicate rapidly improving, but still error-prone, mechanical engineering awareness of current LLMs.
\end{abstract}


\end{frontmatter}


\section{Introduction}
\label{sec:intro}
After early successes in statistical language modeling~\cite{rosenfeld2000twoDecadsOfLanguageModeling}, the introduction of the transformer architecture~\cite{vaswani2017attentionIsAllYouNeed} together with improvements in training data and strategy~\cite{brown2020GPT3, ouyang2022_originalRLHF} enabled modern Large Language Models (LLMs) to excel not only at natural language processing but also at code generation and syntactic reasoning~\cite{zhuo2024bigcodebench, jimenez2024SWEBench, chen2021evaluating}.

Although modern LLMs show remarkable performance on many benchmarks including mathematical reasoning using vision~\cite{wang2024mathVision}, even strong frontier models struggle with (spatial) geometry without reasoning~\cite{Abdullaeva2026NoReGeo}. 
In mechanical engineering, system synthesis requires more than logical syntax and it demands what we define as \textit{Mechanical Engineering Awareness}: the ability to comprehend spatial relationships, exact geometries, physical parameters, and constraints within a MultiBody System (MBS). 

MBSs are a class of mechanical systems and the central subject of this article. In the most general sense, a MBS consists of interconnected bodies that may be rigid or flexible \cite{Woernle2024, Wittenburg2008}, i.e. deformable. This broad definition encompasses a wide spectrum of machines, ranging from academic examples such as pendulums and simple mechanisms, to complex machinery including robots, cranes, and vehicles. MBSs can also represent civil engineering structures. The breadth makes MBSs a compelling testbed for the mechanical engineering awareness of LLMs. 

Producing such mechanical systems from text demands exact, simulation-ready geometry -- a requirement that current generative approaches do not meet. 
Current text-to-3D geometry generation pipelines moved from voxel-based approximations~\cite{sella2023voxE} to Neural Radiance Fields (NeRFs)~\cite{poole2023dreamfusion} and 3D Gaussian Splatting~\cite{tang2024dreamgaussian}, which are optimized for visual aesthetics rather than deterministic engineering constraints. This paper introduces a novel, structured evaluation pipeline capable of generating exact 3D solid geometries and finite element meshes from text, using geometry primitives, making it parametric and bypassing both invalid meshes and voxel hallucinations. By comparing the resulting mechanical system graphs against expert ground-truth models, structural correctness can be verified and differentiated from parameter errors. 
The exact nature of mechanical reasoning errors on problems spanning three different levels of complexity is examined starting from simple rigid to complex flexible MBSs. 

In this study, the mechanical engineering benchmark ``MecEng'' is presented, spanning a wide array of mechanical problems with ground truth implementations for automated evaluation. The MecEng benchmark is evaluated on more than 30 local open-weight LLMs as well as Anthropic's Claude Opus 4.8 and OpenAI's GPT-5.2 via the respective Application Programming Interfaces (APIs)\footnote{Throughout the article, we refer to LLMs by the names used in the Ollama library \url{https://ollama.com/search}, \url{https://github.com/ollama/ollama-python}. The proprietary models are referred to by their name and version: ``Claude-Opus-4.8'' and ``GPT-5.2''.}.

\subsection{State of the Art}
\label{sec:sota}
To generate 3D geometries from text for artistic and graphics applications, methods like DreamFusion~\cite{poole2023dreamfusion}, DreamGaussian~\cite{tang2024dreamgaussian}, Magic3D~\cite{lin2023magic3d}, and Point-E~\cite{nichol2022pointE} use knowledge from pretrained 2D vision-language and text-to-image diffusion models. Using techniques such as CLIP-based guidance or score distillation, these methods either optimize neural scene representations like NeRFs or directly generate point clouds from natural language prompts.
Newer work like MeshGPT~\cite{siddiqui2024meshGPT} and LLaMA-Mesh~\cite{wang2024llamaMesh} demonstrate AI-driven triangle mesh synthesis, but target visual aesthetics rather than the engineering-grade deterministic and watertight geometries required for Finite Element Method (FEM) simulation. 
Thengane et al.\ \cite{Thengane2025PointCloudsSurvey} present a detailed overview of the comprehension of foundational models for 3D point clouds in combination with 2D images. While this shows the abilities of LLMs regarding visually high quality models, in our tests with tools like LLaMA-Mesh and MeshGPT already the generation of simple geometries found in mechanical engineering, such as a cylinder or shouldered shaft, is not appropriate for engineering simulations.
Unlike visual representations for computer graphics, to model flexible MBSs, exact (closed) geometry is required in the form of nodes, meshes and vertices, where the exact geometry is not only relevant for inertial properties, but also defines the body's compliance. 
To achieve this, rather than burdening the model with placing individual vertices and connecting them into meshes, another line of work has LLMs operate through CAD tools instead. Text2CAD~\cite{khan2024text2cad} creates a new DeepCAD dataset using multimodal AI models and fine-tune a pretrained BERT encoder using a custom dataset, while the more recent CAD-coder~\cite{guan2026cad} fine-tunes the Llava1.5 model~\cite{liu2023visualInstructionTuning_LLAVA} to create CAD-code based on Open CASCADE Technology (OCCT) from images. 

Recent work at the intersection of LLMs and engineering simulation motivates our emphasis on zero-shot, local, simulation-grounded model creation and rigorous, automated validation. 
Ni and Buehler~\cite{Ni2024MechAgents} demonstrate that multi-agent LLM systems can formulate and solve mechanics tasks, and that adding specialized critic agents improves error detection in model setup and post-processing. They document failure modes,  e.g., stress component extraction or geometry omissions. Molinari et al.~\cite{molinari2026engiai} present EngiAI, a multi-agent framework for engineering workflows, which also includes a benchmark suite. Notably, their evaluation reports that automatically exported geometry is consistently non-watertight, corroborating our finding that AI-generated geometry is often unsuitable for direct simulation.
For the code-generation of LLMs for finite element tasks at a lower level, like the core element implementation (shape functions, stiffness matrix), Mohammadzadeh et al.\ \cite{Mohammadzadeh2025FEMbench} introduce FEM-Bench. Similarly, Guo et al.\ \cite{Guo2026LLMCAE} utilize LLMs for data-free model order reduction for speeding up simulations for large-scale problems. 
Our work aims to generate simulation models on a higher level, utilizing tested and proven simulation code instead of requiring the model to create code like local stiffness matrices. 
An approach that is closer to our model generation framework is presented by Jadhav and Farimani~\cite{Jadhav2026LLMmechDesigner}, who show that an LLM agent with tool-calling can act as an iterative mechanical designer, however, focusing on iterative refinement (optimization) of structural models rather than CAD-like geometries.

In our recent paper~\cite{Moeltner2026_CreationEvaluationSelfValidation} we propose a parameterized virtual lab for creating, evaluating, and self-validating multibody simulation models with LLMs using in-context learning, including evaluation metrics and automated in-silico experiments. Our recent paper builds the foundation for the LLM evaluation and automated ground-truth comparison via simulation outputs of the present paper. 
EngiBench~\cite{Zhou2025EngiBench} outlines an engineering relevance filter, discipline classification, and a capability-oriented rubric design for evaluating LLMs -- complementary to our evaluation goals, as we similarly need structured rubrics for system assembly, executability, and flexible part correctness. 
Their capability scoring rubric pattern provides a template, which is similar to our graph comparison metric. 
%
Naser et al.\ \cite{Naser2026ERIbenchmark} introduce the Engineering Reasoning and Instruction (ERI) Benchmark, which is a taxonomy-driven benchmark for engineering, using LLM-as-judge evaluation, however, without details on the types of tasks. 
To apply AI methods to engineering problems, Elhambakhsh et al.\ \cite{Elhambakhsh2025LLMmechAssembly} fine-tune LLMs for classifying the function of mechanical assembly from textual component and assembly names, using the Oregon State Design Repository, and report achieving significant accuracy gains compared to general-purpose language models. While solving this problem requires some engineering understanding from the LLM, no spatial information is utilized. 
Guo et al.\ \cite{Guo2025EngDesignBench} benchmark commercial LLMs in various fields of engineering covering more broad disciplines than the present study, however, with tasks that are less specific, like system design with parameter boundaries, and evaluation based on simulations and human evaluation.
%

The spatial reasoning, which is required for the mechanical engineering awareness as well as for robotics, is reviewed by Sharma~\cite{Sharma2023SpatialReasoningLLM}, who identifies fundamental spatial reasoning deficits in LLMs and proposes improvements through instruction tuning, directly motivating our emphasis on spatial geometry and 3D multibody assembly tasks.
The spatial understanding by LLMs is also studied by Yamada et al.\ \cite{Yamada2024SpatialLLM}, where participant/task considerations are highlighted which are relevant to our multi-step spatial/structural reasoning needs in assembling correct simulation geometries and boundary conditions before validation.  
Tang and Kejriwal~\cite{Tang2025GRASP} introduce GRASP, a grid-based commonsense spatial reasoning benchmark, quantifying where LLMs succeed and fail on text-based layout understanding, which is less abstract than the joints and connectors in MBSs in the present study, but require similar spatial reasoning.
In the work of Zeng et al.\ \cite{Zeng2025LEGOpuzzles}, multi-step spatial reasoning is evaluated via LEGO puzzles using a multiple-choice format for automated evaluation. In opposition to the present study, they build mostly on multimodal LLMs.
Tian et al.\ \cite{Tian2024MechEngLLM} examine LLMs' conceptual understanding in mechanical engineering education -- motivating our focus on not just code executability, but conceptual correctness of materials, engineering statics and dynamics, boundary conditions, and post-processing.
The geometry benchmark NoReGeo~\cite{Abdullaeva2026NoReGeo} was introduced by Abdullaeva et al. to evaluate the intrinsic geometric understanding of LLMs without multi-step reasoning scaffolds. While this is similar to our spatial reasoning, in NoReGeo the focus is related to geometrical tasks, such as line-line intersection within an image, while our tests are related to building full mechanical simulations.

These LLM evaluations in the engineering-domain follow three approaches: comparison against human participants or experts~\cite{Tian2024MechEngLLM, Abdullaeva2026NoReGeo, Zeng2025LEGOpuzzles, Han2026LLMtoolSelection, Sharma2023SpatialReasoningLLM}; scoring by LLM judges~\cite{Naser2026ERIbenchmark, Zhou2025EngiBench}; and fully automated evaluation through code execution, simulation-derived ground truth, or rule-based matching~\cite{Moeltner2026_CreationEvaluationSelfValidation, Mohammadzadeh2025FEMbench, Guo2025EngDesignBench, Ni2024MechAgents, Ezemba2025OpenSeeSimE, Jadhav2026LLMmechDesigner, Jiang2026LLMfailureModes, Yamada2024SpatialLLM, Li2026STEMVerse}, with the present work belonging to the last category, comparing LLM-generated multibody simulation models against parameterized ground-truth implementations via system graph isomorphism and numerical time-series validation.

Still, no existing work combines spatial reasoning evaluation, exact geometry generation, and automated engineering validation in a unified benchmark for mechanical systems.

\subsection{Objectives and Novelties}
\label{sec:objectives}

The primary objective of this research is to build a fully automated engineering-grade benchmark for LLMs, the MecEng, which is particularly related to mechanical and multibody models for simulation.
A sub-goal of the paper emerges due to the necessity to compare the LLM-generated mechanical models with the ground truth on different levels, including numerical results, system graph representations, geometries and finite element meshes, requiring exact model definition to achieve uniqueness of representations and solutions.
Another sub-goal is performance evaluation of many open-weight LLMs that run locally on workstations, in particular measuring parameter dependencies such as temperature and chain-of-thought reasoning, also termed thinking. In particular, the dependence of performance on model size and release date shows a continued increase of performance regarding mechanics. This suggests further growth in the future, not only for the field of mechanics, but also for other engineering disciplines.

Novelties include quantifying mechanical engineering awareness, fully automating mechanical model comparison, and providing a generation pipeline from text to finite element models and MBSs, including industry-standard reduced order models.

\section{Mechanical Tasks and Generation Pipeline}
\label{sec_Mechanical Tasks and Generation Pipeline}
The core of the tasks and pipeline for MecEng is based on a previous paper \cite{Moeltner2026_CreationEvaluationSelfValidation}, which has been substantially extended. The set of mechanical tasks has been more than doubled and the difficulty of tasks increased substantially. One of the main reasons for this expansion was that state-of-the-art LLMs have achieved success rates of nearly 100\% on the existing set of tasks.

This section introduces the new task structure and the pipeline, i.e. the process of translating a textual description into a multibody simulation. Furthermore, a dedicated pipeline for MBSs with flexible parts is presented.

\subsection{Mechanical Models and Tasks}
\label{sec_Mechanical Models and Tasks}
In the present paper, we include three difficulty levels, with level 3 split into part generation (3a) and assembly (3b):

\begin{itemize}[align=left, labelwidth=1.5cm, leftmargin=2.25cm]
  \item[{\bf level 1}] simple mechanical models consisting of mass points and single rigid bodies, connected by springs and distance constraints. Loads are available as forces and torques, which may contain user-defined functions over time. Typical examples include single mass oscillators, freely flying rigid bodies, and slider-crank mechanisms, testing the basic multibody modeling, spatial connectivity, and mass/inertia assignments.
  
  \item[{\bf level 2}] medium complex multibody models, including systems of rigid bodies with classical joints such as prismatic, revolute, and rolling joints. Furthermore, advanced models with sphere-sphere and sphere-triangle contact, simple vehicles and scaffolds are added. Typical examples include double pendulums and more complex slider-crank mechanisms, testing the ability to handle increased topological complexity and multi-degree-of-freedom interfaces.
  
  \item[{\bf level 3a}] flexible parts, based on 3D geometries, converted into tetrahedral finite element meshes, with joint interfaces and model order reduction. Typical examples include flexible rotors, flexible bearing blocks, and flexible rods, testing the ability of complex geometry synthesis, assignment of interfaces, and usage of computational tools to model flexible machine parts.
  
  \item[{\bf level 3b}] flexible MBS assemblies, using the parts generated in the level 3a tasks. This level evaluates the correct connection of flexible machine parts at their interfaces.
\end{itemize}

\begin{figure}[htbp]
	\centering
	\begin{subfigure}[b]{0.45\textwidth}
		\centering
		\includegraphics[width=\textwidth]{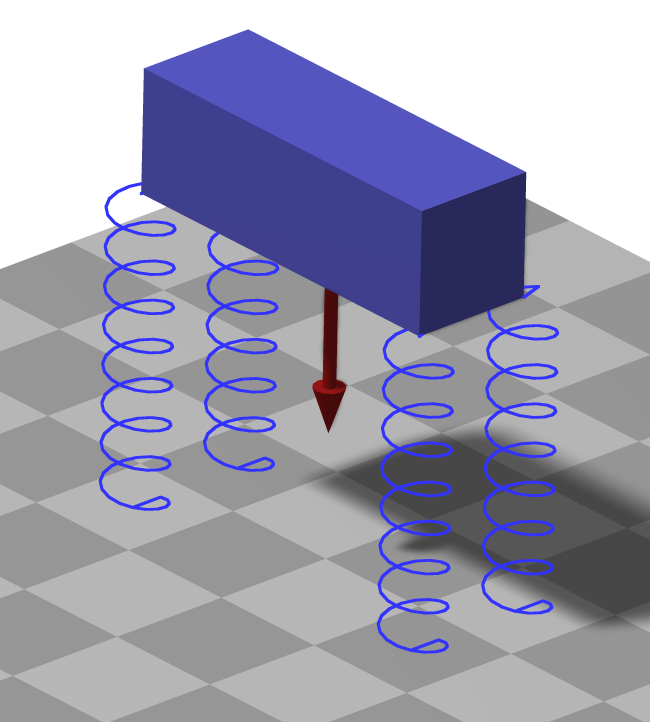}
		\caption{level 1 task}
	\end{subfigure}
	\hfill
	\begin{subfigure}[b]{0.45\textwidth}
		\centering
		\includegraphics[width=\textwidth]{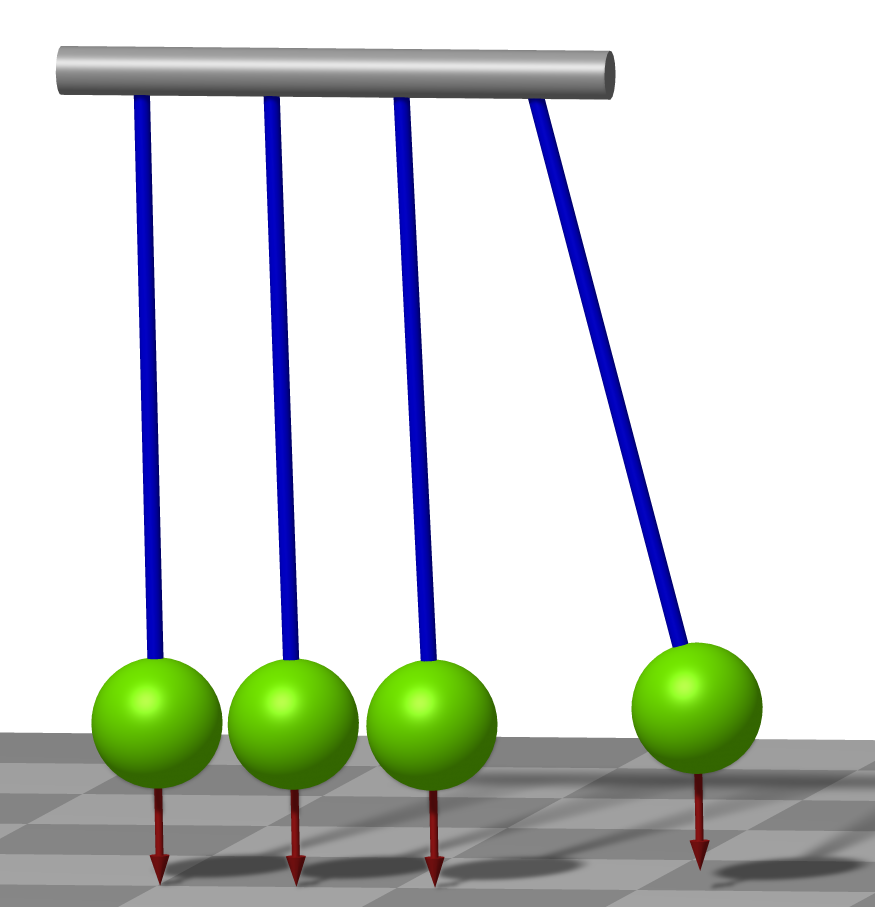}
		\caption{level 2 task}
		\label{fig:level2task}
	\end{subfigure}
	\vspace{0.5em}
	\begin{subfigure}[b]{0.45\textwidth}
		\centering
		\includegraphics[width=\textwidth]{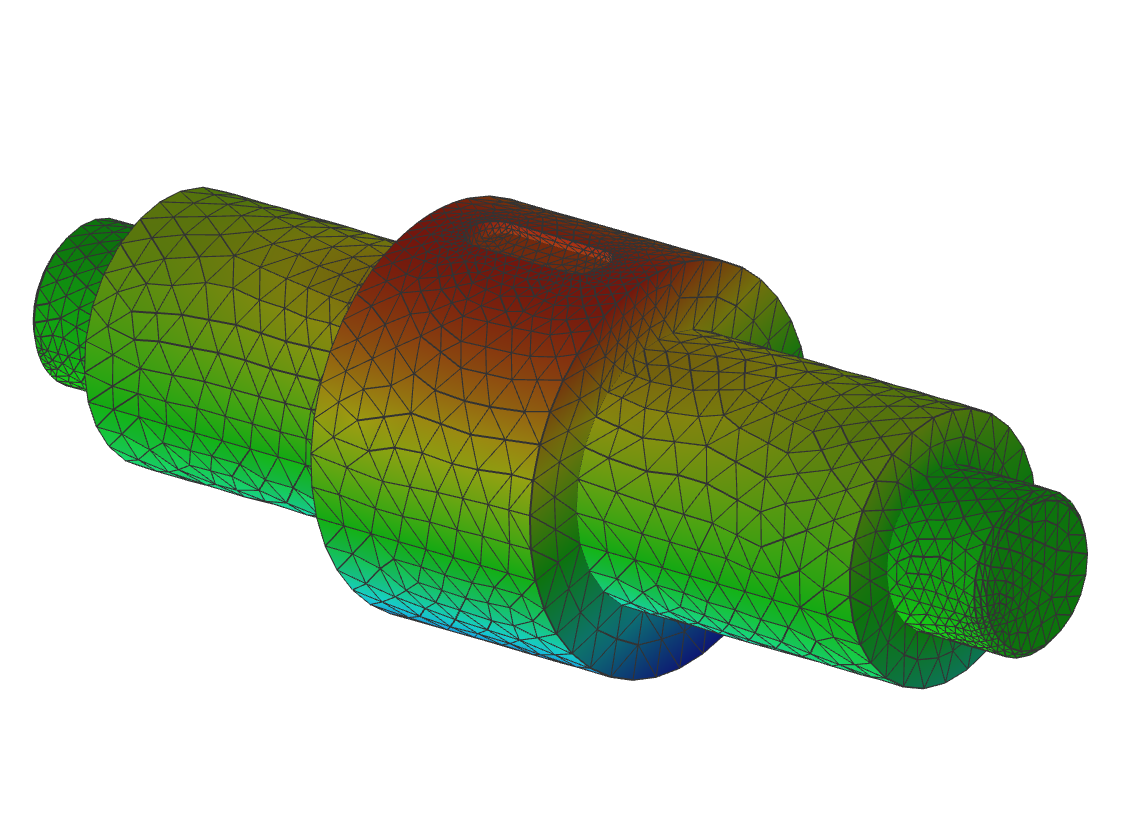}
		\caption{level 3a task}
	\end{subfigure}
	\hfill
	\begin{subfigure}[b]{0.45\textwidth}
		\centering
		\includegraphics[width=\textwidth]{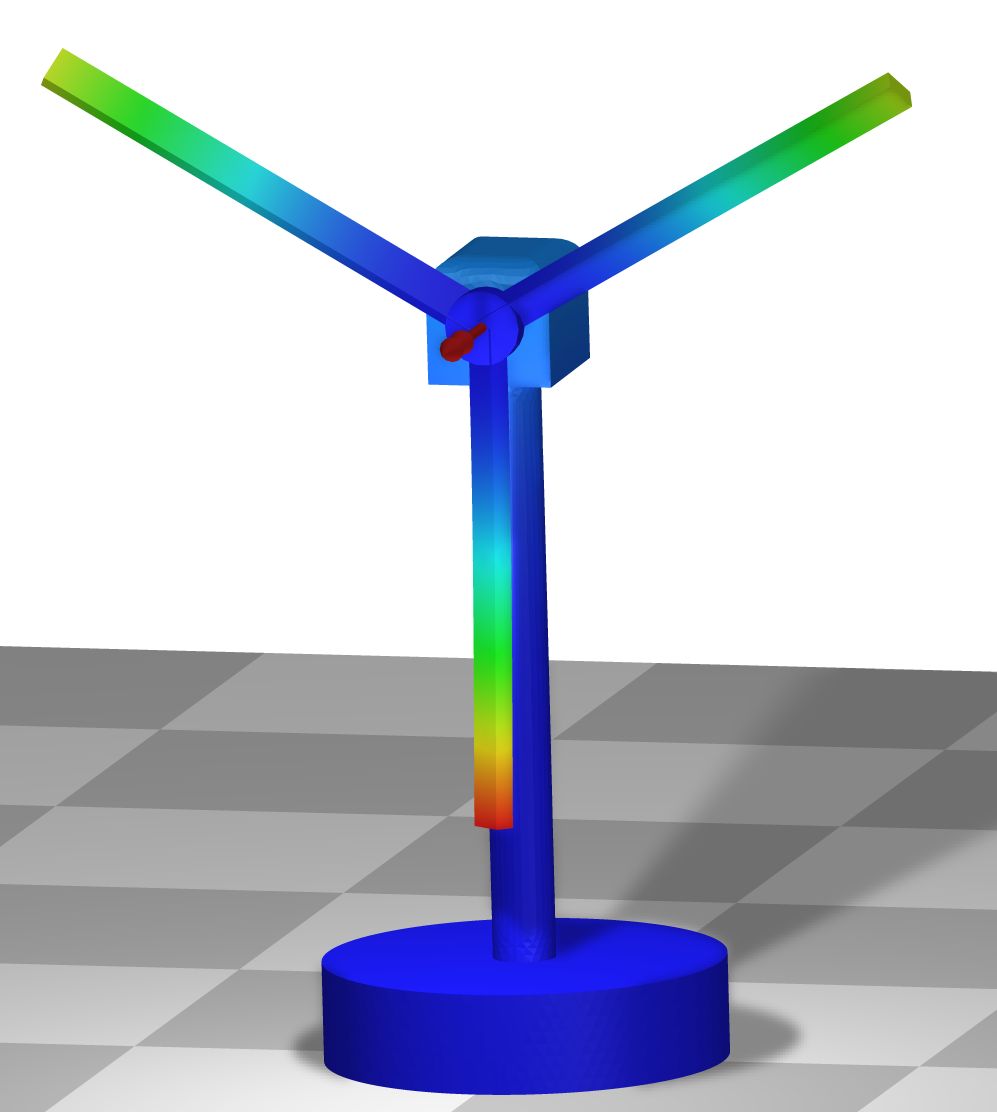}
		\caption{level 3b task}
	\end{subfigure}
	\caption{Solution examples of the four task types investigated in this work. All four examples are fully LLM-generated by the open-weight LLM gemma4:31b. Dimensions and visualization settings were changed manually afterwards for better visibility within the figure.}
	\label{fig:resultsExamples}
\end{figure}

In \fig{fig:resultsExamples}, three LLM-generated MBSs and one flexible machine part are shown. In (a), the LLM was instructed to create a simulation for a ``suspended rigid body''. The rigid body is connected to ground using four spring-dampers shown in blue. 
In (b), a ``Newton's cradle'' is simulated. The LLM is required to create the mass points, the distance connectors shown in blue, and a sphere-sphere contact between the spheres. In (c), the ``balanced rotor part'' generated by the LLM is shown. The rotor is a stepped shaft with three distinct diameters. Chamfers are added at both ends of the shaft and a keyway sits in the middle section, generated by extruding the base face of the keyway profile and subtracting it from the shaft body. The color indicates the normalized displacement of the second eigenmode. Note that two interfaces are added by the LLM, one on the left and one on the right side; using these interfaces, the part can subsequently be included in a level 3b task. In (d), the flexible MBS ``simplified wind turbine'' is shown. The LLM assembles the successfully created ``wind turbine rotor part'' and ``wind turbine tower part''. The color grading indicates the resulting local displacement due to the torque applied around the rotor axis.

In total, there are 30 level 1 tasks, 20 level 2 tasks and 34 level 3 tasks, see \mytab{tab:modelList}. All tasks are widely parameterized. Using randomized parameters, it is possible to create thousands of unique tasks per mechanical model for statistical evaluation. Note that parameterization may also encompass structural variations such as differing numbers of bodies, coordinate system alignments, or interchangeable parts serving the same role within a given flexible MBS. For example, the ``flexible pendulum'' MBS randomly requires one out of several arm geometries: a simple flexible beam, a beam with fillets, a beam with a hole, or a thin-walled profile.

Regardless of the level, mechanical models are always provided in generic form as text templates, with the model generator typically inserting 5 to 10 parameters, such as mass, gravity, dimensions, and stiffness, into the template. A typical level 2 model template is as follows, where the resulting MBS generated by the LLM is already shown in \fig{fig:resultsExamples}(b).

\begin{quote}
  \footnotesize
  \textbf{Newton's cradle:}
  \textit{A Newton's cradle consisting of 4 identical spheres (index 0 to 3) modeled as mass points. Each sphere has a mass m = \{mass\} kg, a radius r = \{radius\} m, and is suspended from a fixed ceiling (ground) at z = \{length\} m. Each sphere with index i is connected to the ceiling at p\_fix,i = $[$i*(2*\{radius\}+0.002), 0, \{length\}$]$ via a distance constraint of length L = \{length\} m. The spheres are aligned along the x-axis. Spheres 1, 2, and 3 are initially at rest at their equilibrium positions p\_i = $[$i*(2*\{radius\}+0.002), 0, 0$]$. Sphere 0 is initially positioned at p\_0 = $[$-\{length\}, 0, \{length\}$]$, representing a horizontal configuration with the cable taut, and starts from rest. Gravity g = \{gravity\} m/s\textasciicircum{}2 acts in the negative z-direction. Contact between spheres is modeled with a stiffness k = \{stiffness\} N/m and a restitution coefficient e = \{restitution\}. Friction is neglected. Add the spheres in the order of their index.}
\end{quote}

In the end, every task requires to build a simulation model in Exudyn\footnote{\url{https://github.com/jgerstmayr/EXUDYN}}~\cite{Gerstmayr2024} for the given parameterized mechanical model. Exudyn is a multibody simulation code that is used via a Python interface. Thus, the core objective of the pipeline is to transform a text-based engineering description into a functional, mathematically accurate MBS simulation model, given as a self-contained Python code.

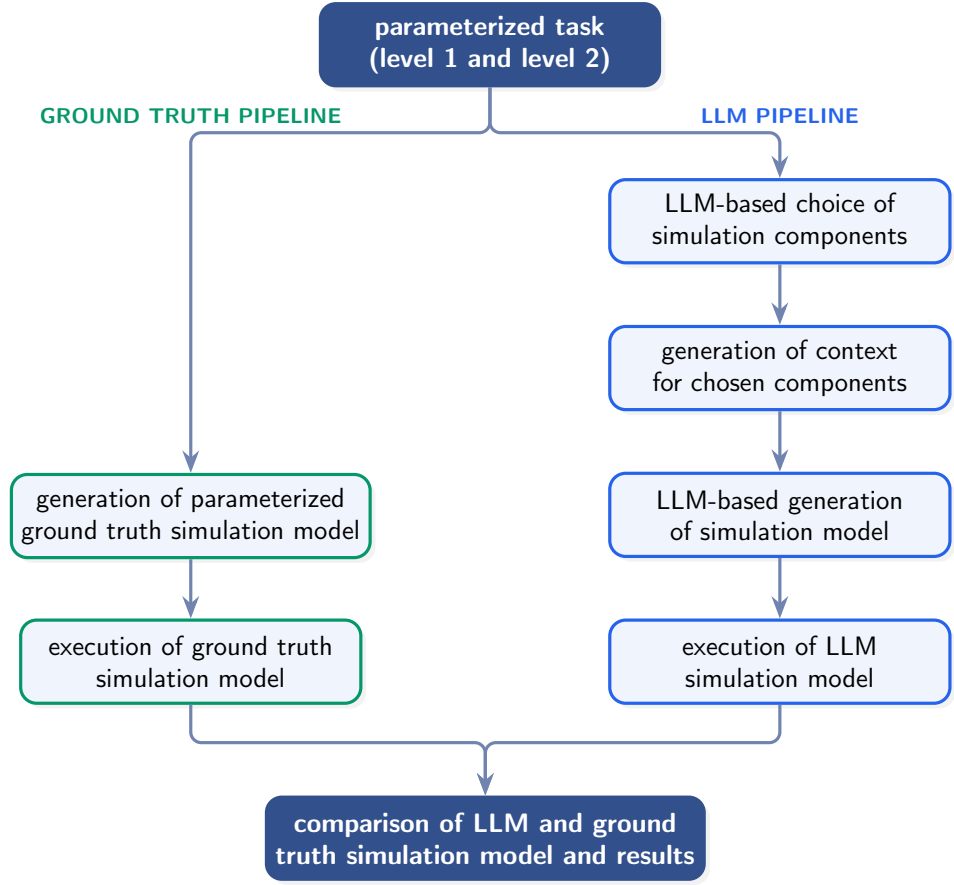
\begin{figure}[tbh]
	\centering
	\begin{tikzpicture}[
		node distance = 0.8cm and 1.5cm,
		>=Stealth,
		block/.style = {
			rectangle, 
			draw=primary!20, 
			fill=bglight, 
			minimum width=4.5cm, 
			minimum height=1.1cm, 
			align=center, 
			rounded corners=6pt,
			line width=0.8pt,
			font=\sffamily\small,
			drop shadow={opacity=0.1, shadow xshift=2pt, shadow yshift=-2pt}
		},
		startend/.style = {block, draw=primary, fill=primary, text=white, font=\sffamily\bfseries\small},
		llm/.style = {block, draw=llmbeam, line width=1.2pt},
		gt/.style = {block, draw=gtbeam, line width=1.2pt},
		line/.style = {draw=primary!70, ->, line width=1.2pt, rounded corners=4pt}
		]
		
		
		\node [startend] (step1) {parameterized task\\(level 1 and level 2)};
		
		\node [llm, below right = 1.2cm and -0.7cm of step1] (step2a) {LLM-based choice of\\simulation components};
		\node [gt, below left = 5.1cm and -0.7cm of step1] (step2b) {generation of parameterized\\ground truth simulation model};
		
		\node [llm, below = of step2a] (step3a) {generation of context\\for chosen components};
		\node [llm, below = of step3a] (step4a) {LLM-based generation\\of simulation model};
		\node [llm, below = of step4a] (step5a) {execution of LLM\\simulation model};
		
		\node [gt] (step3b) at (step5a -| step2b) {execution of ground truth\\simulation model};
		
		\coordinate (mergePoint) at ($(step5a.south)!0.5!(step3b.south)$);
		\node [startend, below = 1.2cm of mergePoint] (step6) {comparison of LLM and ground\\truth simulation model and results};
		
		
		\path [line] (step1.south) -- ++(0,-0.6) -| (step2a.north);
		\path [line] (step1.south) -- ++(0,-0.6) -| (step2b.north);
		
		\path [line] (step2a) -- (step3a);
		\path [line] (step3a) -- (step4a);
		\path [line] (step4a) -- (step5a);
		
		\path [line] (step2b) -- (step3b);
		
		\path [line] (step5a.south) |- ($(step6.north) + (0, 0.7)$) -- (step6.north);
		\path [line] (step3b.south) |- ($(step6.north) + (0, 0.7)$) -- (step6.north);
		
		\node[text=llmbeam, font=\sffamily\bfseries\scriptsize, above=0.6cm of step2a] {LLM PIPELINE};
		\node[text=gtbeam, font=\sffamily\bfseries\scriptsize, above=4.5cm of step2b] {GROUND TRUTH PIPELINE}; 
		
	\end{tikzpicture}
	\caption{Workflow for generation of simulation models using expert (ground truth) and LLM implementations. At the beginning, a task is selected from the available tasks (depending on the investigated level) and randomized parameters are inserted.}
	\label{fig:workflowGeneration}
\end{figure}

\subsection{Simulation Model Workflow}
\label{sec_Simulation Model Workflow}
The workflow for generating simulation models is sketched in \fig{fig:workflowGeneration}, which is done for each task. Note that for level 3 tasks, a modified version of this workflow is applied, which is described later.

A crucial part of the LLM pipeline is the context generation. As we assume that most LLMs have not been fully trained and finetuned on our simulation environment Exudyn, we provide specialized context which is provided together with the task. In a pre-step, the LLM has to choose from a list of simulation components (25 in the case of level 2 tasks). This is similar to typical Retrieval-Augmented Generation~(RAG)-approaches~\cite{BaumannEberhard2026}, however, giving us the ability to exactly track the chosen components and to provide 25 components within only \num{19825} characters for medium MBSs, if all tags are chosen. This would require a context size of approx.\ 5\,000 tokens solely for the components information, however, being much shorter in the tests as only a maximum of six different components is required for the most complicated examples. The context is then generated based on the chosen simulation components, noting that the LLM mostly fails if essential components are not chosen -- except that a special component like a joint adds dependent components like ground or a rigid body. 

Based on the context and the model description, the final task prompt is generated for the LLM. The main task herein is the generation of a Python code for the simulation model. Details of LLM inference with different LLM interfaces are shown in \refSection{sec_LLMinference}. At the same time, a parameterized ground truth simulation model is generated based on the same randomized parameters. In the final step, both simulation models are executed independently and checked for errors. Several metrics are used to compare the models and numerical results, as shown in detail in \refSection{sec_modelComparison}.

For the explained workflow, different tracking and logging procedures are set up, which are explained in the next subsections.

\subsubsection{Configuration and Settings}
\label{sec_Configuration and Settings}
Each experimental run is governed by settings organized into three groups. The \emph{LLM settings} specify which model is used, the inference backend, quantization, context window size, generation temperature, and whether extended chain-of-thought reasoning is active. The \emph{LLM pipeline settings} control the scope of the benchmark, in particular the number of random parameter variations generated per task model, context size and maximum tokens for each pipeline stage, the solver timeout, and the numerical tolerance used for solution comparison. A third group of \emph{prompt settings} selects among a set of interchangeable prompt templates for both the simulation-component selection step and the code generation step, enabling ablation studies on prompt design without any other change to the pipeline.

All settings, together with GPU hardware telemetry and the Exudyn simulator version, are written to a single machine-readable record at the end of each run. This record also accumulates the aggregated evaluation results, so that the complete experimental context is captured in one file per run.

\subsubsection{Task-Level Tracking}
\label{sec_Task-Level Tracking}
For every individual task execution, the pipeline tracks two stages of data.
During the first \emph{context selection stage}, the full prompt sent to the LLM is stored alongside the model's response and a record of which simulation components were selected, including any components that are surplus to or missing from the ground-truth reference set. During the second \emph{code generation and evaluation stage}, the prompt and the raw LLM response are retained together with the extracted simulation code, the file paths of the generated and reference source files, and the complete outcome of every evaluation metric described in \refSection{sec_modelComparison}.

This task-level record enables manual inspection of individual tasks and, once aggregated across all task runs, yields the success counts and rates used in the statistical evaluation and for the results presented in this article.

\subsubsection{Logging and Outputs}
\label{sec_Logging and Outputs}
The pipeline produces a self-contained JSON file per run, capturing the full configuration, hardware telemetry, aggregated metrics, and all per-task records -- everything needed to reproduce the reported statistics. Narrative and debug logs complement the JSON file with prompts, LLM responses, timing, token counts, and automatic error/warning summaries. Importantly, LLM-generated and reference Python source files are retained so that any simulation can be re-executed independently of the pipeline. Each file embeds its parameterized task description as a header comment, yielding thousands of self-contained, directly executable description–code pairs that are straightforward to inspect and verify.

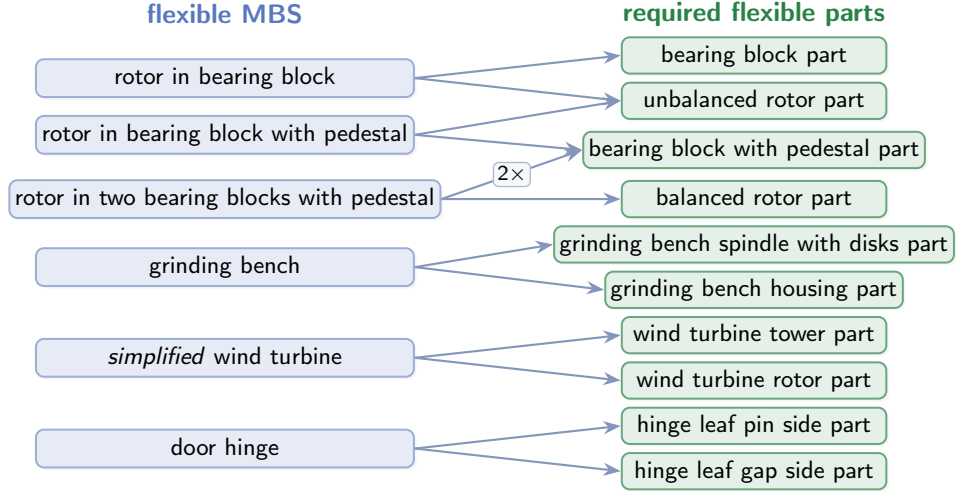
\begin{figure}
	\centering
	\begin{tikzpicture}[
		>=Stealth,
		mbs/.style = {
			rectangle,
			draw=mbsblue!60, fill=mbsblue!15,
			minimum width=5cm, minimum height=0.50cm,
			align=center, rounded corners=3pt,
			line width=0.75pt,
			font=\sffamily\footnotesize,
			inner sep=2pt,
			drop shadow={opacity=0.08, shadow xshift=1pt, shadow yshift=-1pt}
		},
		fixedpart/.style = {
			rectangle,
			draw=fixedgreen!70, fill=fixedgreen!12,
			minimum width=3.5cm, minimum height=0.48cm,
			align=center, rounded corners=3pt,
			line width=0.75pt,
			font=\sffamily\footnotesize,
			inner sep=2pt,
			drop shadow={opacity=0.08, shadow xshift=1pt, shadow yshift=-1pt}
		},
		solidline/.style = {draw=primary!60, ->, line width=0.8pt},
		header/.style = {font=\sffamily\bfseries\small, align=center}
		]
		\node[header, text=mbsblue]   (hMBS) at (0,   0) {flexible MBS};
		\node[header, text=fixedgreen](hFix) at (7, 0) {required flexible parts};
		\node[mbs] (mbs1) at (0, -0.85) {rotor in bearing block};
		\node[fixedpart] (fp1a) at (7, {-0.85+0.30}) {bearing block part};
		\node[fixedpart] (fp1b) at (7, {-0.85-0.30}) {unbalanced rotor part};
		\path[solidline] (mbs1.east) -- (fp1a.west);
		\path[solidline] (mbs1.east) -- (fp1b.west);
		\node[mbs] (mbs2) at (0, -1.60) {rotor in bearing block with pedestal};
		\node[fixedpart] (fp2a) at (7, -1.80) {bearing block with pedestal part};
		\path[solidline] (mbs2.east) -- (fp2a.west);
		\path[solidline] (mbs2.east) -- (fp1b.west);
		\node[mbs] (mbs3) at (0, -2.45) {rotor in two bearing blocks with pedestal};
		\node[fixedpart] (fp3a) at (7, -2.45) {balanced rotor part};
		\path[solidline] (mbs3.east) -- (fp3a.west);
		\path[solidline] (mbs3.east) -- node[font=\scriptsize\sffamily, draw=primary!40, fill=mbsblue!5, rounded corners=2pt, inner sep=1.5pt, line width=0.5pt] {2×} (fp2a.west);
		\node[mbs] (mbs4) at (0, -3.35) {grinding bench};
		\node[fixedpart] (fp4a) at (7, {-3.35+0.30}) {grinding bench spindle with disks part};
		\node[fixedpart] (fp4b) at (7, {-3.35-0.30}) {grinding bench housing part};
		\path[solidline] (mbs4.east) -- (fp4a.west);
		\path[solidline] (mbs4.east) -- (fp4b.west);
		\node[mbs] (mbs5) at (0, -4.55) {\emph{simplified} wind turbine};
		\node[fixedpart] (fp5a) at (7, {-4.55+0.30}) {wind turbine tower part};
		\node[fixedpart] (fp5b) at (7, {-4.55-0.30}) {wind turbine rotor part};
		\path[solidline] (mbs5.east) -- (fp5a.west);
		\path[solidline] (mbs5.east) -- (fp5b.west);
		\node[mbs] (mbs6) at (0, -5.75) {door hinge};
		\node[fixedpart] (fp6a) at (7, {-5.75+0.30}) {hinge leaf pin side part};
		\node[fixedpart] (fp6b) at (7, {-5.75-0.30}) {hinge leaf gap side part};
		\path[solidline] (mbs6.east) -- (fp6a.west);
		\path[solidline] (mbs6.east) -- (fp6b.west);
	\end{tikzpicture}
	\caption{Overview of the flexible MBSs that need two flexible parts each. Be aware that the MBS ``rotor in two bearing blocks'' needs three parts, although two parts of the same type, which is why it is also listed here. In the present work, also flexible MBSs that use one flexible body or three flexible bodies are considered.}
	\label{fig:flexuAIModels2Part}
\end{figure}

\subsection{Flexible MBS}
\label{sec_FlexuAI}
Level 3 flexible MBS tasks require a special workflow that differs from the one shown in \fig{fig:workflowGeneration}. The flexible MBS workflow begins with the selection of a flexible MBS, e.g., ``\emph{simplified} wind turbine'' shown in \fig{fig:resultsExamples}(d). Before addressing the assembly within the level 3b task itself, the workflow loads the descriptions of the required level 3a tasks, e.g., the flexible ``wind turbine rotor part'' and ``wind turbine tower part''. In \fig{fig:flexuAIModels2Part}, some flexible MBSs and the required flexible parts are illustrated. As can be seen, the examples are chosen such that some machine parts are re-used across various flexible MBSs. The pipeline for generating LLM-produced meshed parts is then executed, and only once all parts are evaluated as correct, the workflow proceeds to solve the assembly task.

\begin{figure}[tbh]
	\centering
	\pgfdeclarelayer{behind}
	\pgfsetlayers{background, behind, main}
	\begin{tikzpicture}[
		node distance = 0.8cm and 1.5cm,
		>=Stealth,
		block/.style = {rectangle, 
			draw=primary!20, 
			fill=bglight, 
			minimum width=2cm, 
			minimum height=1.1cm, 
			align=center, 
			rounded corners=6pt,
			line width=0.8pt,
			font=\sffamily\small,
			drop shadow={opacity=0.1, shadow xshift=2pt, shadow yshift=-2pt}
		},
		blockSmall/.style = {rectangle, 
			draw=primary!20, 
			fill=bglight, 
			minimum width=2cm, 
			minimum height=1.1cm, 
			align=center, 
			rounded corners=6pt,
			line width=0.8pt,
			font=\sffamily\small,
			drop shadow={opacity=0.1, shadow xshift=2pt, shadow yshift=-2pt}
		},
		startend/.style = {block, draw=primary, fill=primary, text=white, font=\sffamily\bfseries\small},
		llm/.style = {block, draw=llmbeam, line width=1.2pt},
		gt/.style = {block, draw=gtbeam, line width=1.2pt},
		parameter/.style = {blockSmall, draw=black, line width=1.2pt},
		description/.style = {blockSmall, draw=black, line width=1.2pt},
		line/.style = {draw=primary!70, ->, line width=1.2pt, rounded corners=4pt}]
		\node [startend] (head){parameterized task\\(level 3b)};
		\matrix (subnodes) [
		below=0.3cm of head,
		nodes={align=center, minimum height=1.2cm, anchor=center},
		matrix of nodes,
		column sep=1cm
		] {|(part1Description)[description]| \shortstack{part 1\\description} &
			|(masterParameters)[parameter]| \shortstack{master\\parameters} &
			|(part2Description)[description]| \shortstack{part 2\\description}\\};
		\node [below=1.3 of masterParameters.south](partCreation) {part generation};
		\node[description, below=0.1cm of partCreation] (chooseItemsPart) {LLM-based choice of part generation components};
		\node[description, below=0.3cm of chooseItemsPart] (generateCodePart) {LLM-based generation of part};
		\begin{pgfonlayer}{behind}
			\node[draw, dashed, rounded corners, fill=white, fit=(partCreation)(chooseItemsPart)(generateCodePart), xshift=0.3cm, yshift=0.3cm](partCreationBox1){};
			\node[draw, dashed, rounded corners, fill=white, fit=(partCreation)(chooseItemsPart)(generateCodePart)](partCreationBox){};
		\end{pgfonlayer}
		\node[description, below=0.6cm of generateCodePart] (executePart) {meshing of LLM generated and ground truth \emph{part}};
		\node [startend, below=0.4cm of executePart] (partEvaluation) {comparison of LLM and\\ground truth \emph{part}};
		\node[description, right=1.5cm of partEvaluation] (femObject) {save part};
		\coordinate (pMaster) at (masterParameters.north |- head.south);
		\draw [line] (pMaster) -- (masterParameters.north);
		\draw [line] (head.west) -| (part1Description.north);
		\draw [line] (head.east) -| (part2Description.north);
		\draw [line] (masterParameters.west) -- (part1Description.east);
		\draw [line] (masterParameters.east) -- (part2Description.west);
		\coordinate (mergeLevel) at ($(masterParameters.south)+(0,-0.3cm)$);
		\draw [line] (part1Description.south) |- (mergeLevel) -- (partCreationBox);
		\coordinate (abovePartCreation1) at ($(mergeLevel)+(0.3cm, 0)$);
		\draw [line] (part2Description.south) |- (abovePartCreation1) -- (partCreationBox1);
		\draw [line] (partCreationBox.south) -- (executePart);
		\draw [line] (executePart) -- (partEvaluation);
		\draw [line] (partEvaluation) -- node[above] {success} (femObject);
	\end{tikzpicture}
	\caption{Schematic overview of the first phase of the flexible MBS workflow that is used to generate flexible parts. However, at the beginning of this stage, level 3b tasks determine the required flexible parts and thus the descriptions and master parameters used within the stage. For illustrative purposes, a flexible MBS consisting of two parts is shown.}
	\label{fig:flexuAIPipelineOverviewParts}
\end{figure}
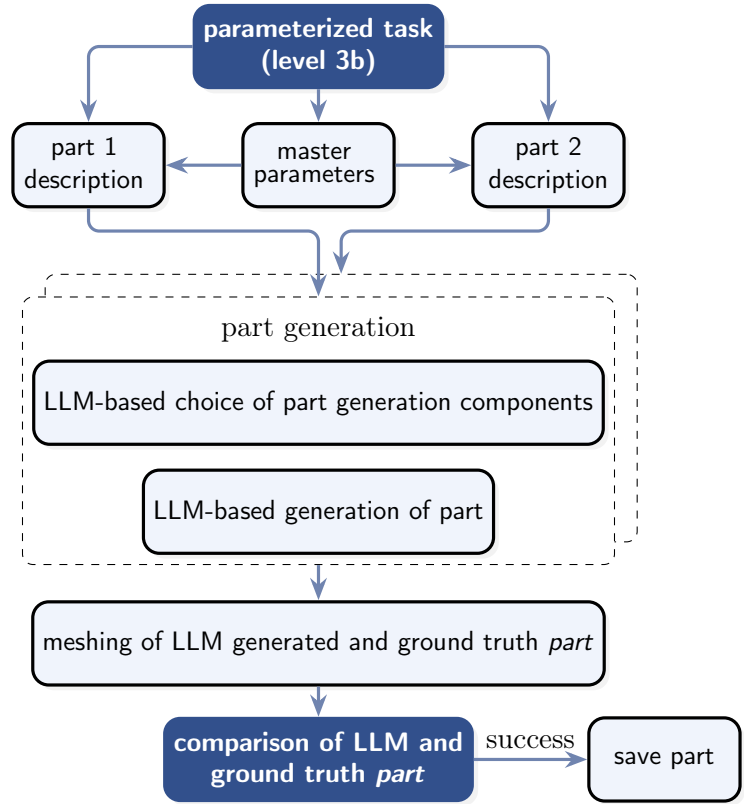

\subsubsection{Part Generation}
\label{Part Generation}
In \fig{fig:flexuAIPipelineOverviewParts}, the part generation phase of the flexible MBS workflow is shown. At the beginning of the part generation phase, two types of information are available:
\begin{enumerate}
	\item Master parameters, that are randomly selected within predefined ranges. The master parameters are shared between the machine and its machine parts. Consider the wind turbine example: the nacelle bore within the ``wind turbine tower part'' must have the same radius as the shaft within the ``wind turbine rotor part''.
	\item Machine part descriptions encode all geometric information about a part in a comprehensive and unambiguous way. Like shown in \refSection{sec_Mechanical Tasks and Generation Pipeline} for Newton's cradle, the machine part descriptions follow a template structure in which some parameters are randomized within a predefined range while others are governed by the master parameters. In addition to the geometric information about a part, the description also holds information about the material of the part and the interfaces that should be defined and named. The interfaces are later used within the assembly task.
\end{enumerate}
Using the retrieved information, similar to the workflow shown in \fig{fig:workflowGeneration} for level 1 and level 2 tasks, the part generation phase in turn consists of a component selection stage and generation stage. The two stages are performed subsequently for each part to be generated.

In the component selection stage, the LLM must select components out of two groups:
\begin{enumerate}
	\item Components to create the geometry of the machine part using a Python wrapper around the OCCT provided by Netgen/NGSolve\footnote{\url{https://github.com/ngsolve/ngsolve}}. Those components contain primitive geometries, the extrusion of an arbitrary face, rotations and translations of created geometries, as well as rounding and bevelling of edges and faces.
	\item Components to transfer the geometry into a finite element mesh, assign materials and interfaces, import it into an interface of our simulator Exudyn, and reduce it using Hurty-Craig-Bampton reduction with specified interfaces, see~\cite{Hurty1965, Craig1968} for theoretical backgrounds.
\end{enumerate}
Once the components have been selected, any surplus or missing components are tracked, and the LLM generates Python code that, when executed, ideally produces a meshed machine part with addressable interfaces, ready for loading into Exudyn's FEM interface.

Before executing the LLM-generated code, we first run the expert reference code to obtain the ground truth part. A pre-execution check is then applied to the LLM code, since mesh parameters, particularly mesh size, can cause excessive runtimes. A minimum mesh size is computed from the ground truth part's bounding box: if the LLM selects a finer mesh, it is silently overridden without penalty, since an overly fine mesh is impractical rather than incorrect. If the mesh is too coarse and the resulting geometry is inaccurate, however, the evaluation flags it as incorrect. This keeps mesh generation runtimes within a reasonable range, as shown in the results section. Note that the LLM is explicitly instructed to hard-code a mesh size (e.g., \texttt{GenerateMesh(maxh=0.01)}), since otherwise the mesh size cannot be extracted out of the code prior to execution, risking excessive runtimes. If an LLM does not follow this rule, the part is flagged, not executed, and rated as incorrect.

After the part generation phase, each part is evaluated using the ground truth part. If the part is evaluated as correct, it is saved using a \texttt{.npz} archive. Thus, the saved parts are available for the flexible MBS assembly phase of the workflow.

\begin{figure}[tbh]
	\centering
	\pgfdeclarelayer{behind}
	\pgfsetlayers{background, behind, main}
	\begin{tikzpicture}[
		node distance = 0.8cm and 1.5cm,
		>=Stealth,
		block/.style = {rectangle, 
			draw=primary!20, 
			fill=bglight, 
			minimum width=2cm, 
			minimum height=1.1cm, 
			align=center, 
			rounded corners=6pt,
			line width=0.8pt,
			font=\sffamily\small,
			drop shadow={opacity=0.1, shadow xshift=2pt, shadow yshift=-2pt}
		},
		blockSmall/.style = {rectangle, 
			draw=primary!20, 
			fill=bglight, 
			minimum width=2cm, 
			minimum height=1.1cm, 
			align=center, 
			rounded corners=6pt,
			line width=0.8pt,
			font=\sffamily\small,
			drop shadow={opacity=0.1, shadow xshift=2pt, shadow yshift=-2pt}
		},
		startend/.style = {block, draw=primary, fill=primary, text=white, font=\sffamily\bfseries\small},
		llm/.style = {block, draw=llmbeam, line width=1.2pt},
		gt/.style = {block, draw=gtbeam, line width=1.2pt},
		parameter/.style = {blockSmall, draw=black, line width=1.2pt},
		description/.style = {blockSmall, draw=black, line width=1.2pt},
		line/.style = {draw=primary!70, ->, line width=1.2pt, rounded corners=4pt}]
		\node [startend] (head){parameterized task\\(level 3b)};
		\matrix (subnodes) [
		below=0.3cm of head,
		nodes={align=center, minimum height=1.2cm, anchor=center},
		matrix of nodes,
		column sep=1cm
		] {|(part1Description)[description]| \shortstack{part 1\\description} &
			|(fMBSDescription)[parameter]| \shortstack{flexible MBS\\description} &
			|(part2Description)[description]| \shortstack{part 2\\description}\\};
		\node [below=0.6 of fMBSDescription.south](fMBSCreation) {flexible MBS generation};
		\node[description, below=0.1cm of fMBSCreation] (chooseItemsFMBS) {LLM-based choice of simulation components};
		\node[description, below=0.3cm of chooseItemsFMBS] (generateCodeFMBS) {LLM-based generation of simulation model};
		\begin{pgfonlayer}{behind}
			\node[draw, dashed, rounded corners, fill=white, fit=(fMBSCreation)(chooseItemsFMBS)(generateCodeFMBS)](fMBSCreationBox){};
		\end{pgfonlayer}
		\node[description, below=0.6cm of generateCodeFMBS] (executeFMBS) {execution of LLM and ground truth simulation model};
		\node [startend, below=0.4cm of executeFMBS] (fMBSEvaluation) {comparison of LLM and ground\\truth simulation model and results};
		\node [description, left=0.6cm of executeFMBS] (savedParts) {saved parts};
		\coordinate (pMaster) at (fMBSDescription.north |- head.south);
		\draw [line] (pMaster) -- (fMBSDescription.north);
		\draw [line] (head.west) -| (part1Description.north);
		\draw [line] (head.east) -| (part2Description.north);
		\draw [line] (part1Description.east) -- (fMBSDescription.west);
		\draw [line] (part2Description.west) -- (fMBSDescription.east);
		\draw [line] (fMBSDescription.south) -- (fMBSCreationBox.north);
		\draw [line] (fMBSCreationBox.south) -- (executeFMBS);
		\draw [line] (executeFMBS) -- (fMBSEvaluation);
		\draw [line] (savedParts) -- (executeFMBS);
	\end{tikzpicture}
	\caption{Schematic overview of the second phase of the flexible MBS workflow that is used to generate the flexible MBSs. The parts saved during the first phase of the workflow are loaded.}
	\label{fig:flexuAIPipelineOverviewFMBS}
\end{figure}
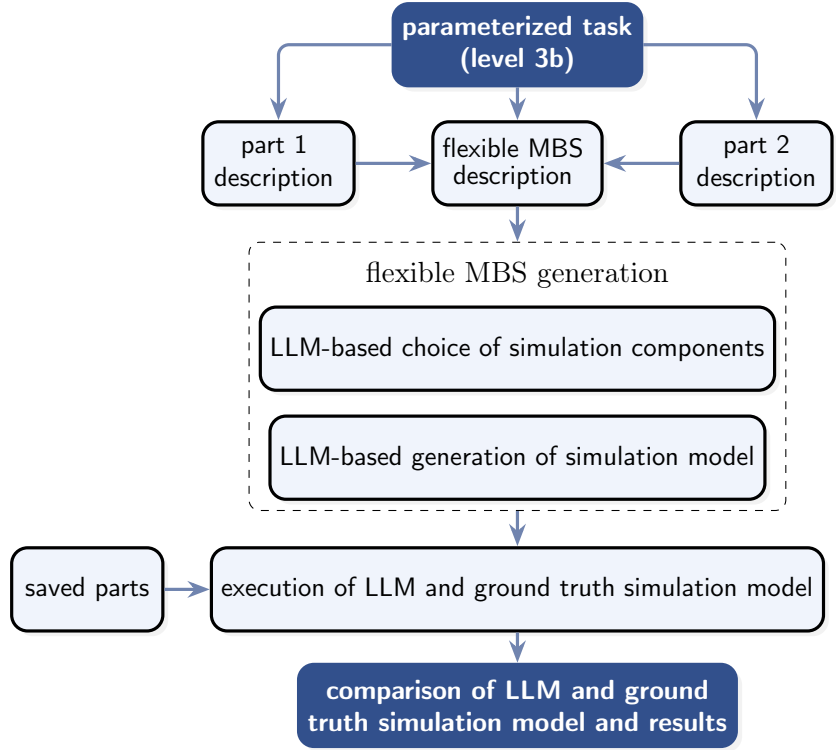

\subsubsection{Flexible MBS Assembly}
\label{sec_Flexible Multibody Systems Assembly}
In \fig{fig:flexuAIPipelineOverviewFMBS}, the flexible MBS assembly pipeline is shown. Starting from the parameterized level 3b task, the pipeline retrieves the flexible MBS description, with randomized parameters filled in, along with the descriptions of the parts generated in the previous part generation phase. Only level 3b tasks whose required parts were evaluated as correct are considered here. If not all parts are available for the level 3b task, it is skipped and rated as not successful. The retrieved part descriptions are appended to the MBS description under the following note:
\begin{quote}
	\footnotesize
	\textbf{Part Loading Info:}\\
	\textit{You already have the \{numberOfPartsAsString\} needed machine part(s) available as a Finite Element Method (FEM) model. Load the corresponding FEM interface(s) with their/its according name(s)!}
\end{quote}
This informs the LLM not only which parts are available, but also their geometries, interfaces, and the names by which they can be loaded.

The generation phase then follows the same workflow as for level 1, level 2, see also \fig{fig:workflowGeneration}: simulation components are selected and an Exudyn simulation is generated. Importantly, the simulation components available to the LLM include a high-level function newly introduced to Exudyn that simplifies the use of flexible bodies in a MBS via the floating frame of reference formulation and model order reduction, see~\cite{Zwoelfer2020} for theoretical background information. This function automatically places markers at predefined, in the part generation stage named interfaces, allowing the LLM to connect parts and apply loads as done within rigid body models.

Once assembly is complete, both the LLM and ground truth simulation models are executed, with both loading the previously saved, LLM-generated part that passed evaluation. Using the same part for both models ensures that the resulting flexible MBS is entirely LLM-generated and that any difference in the evaluation outcome is attributable solely to the assembly stage rather than to discrepancies in the underlying parts.

\subsection{LLM Inference}
\label{sec_LLMinference}
To ensure privacy and data security as well as reproducibility, this work relies mostly on open-weights models, therefore relying on the widely used LLM code Ollama with its associated Python package. In Ollama, models are pulled natively through the command line via \texttt{``ollama pull <modelname>''} and executed entirely locally on workstations. 
By default, this setup utilizes a 4-bit quantization, which is an excellent compromise between generation quality, VRAM requirements, and inference speed. 

The LLM workflow begins with the generation and aggregation of all raw input prompts designed for batched processing. These prompts are passed through the LLMs' standardized prompt templates. The pipeline then initiates batched LLM inference, manually adjusting batch sizes to match hardware VRAM limits and to saturate the threshold beyond which no hardware acceleration yields further speed benefits.
We mention batch-related speedup factors between two and five for our tested LLMs and hardware.
Throughout execution, the backend actively monitors metrics including input and response token counts, overall timing, and execution faults.

In particular we optimize the LLMs' context window size (base value: 4096) as well as the maximum tokens generated (base value: 2048) in order to limit GPU memory requirements and to improve batch-performance. The base values of both token limits are multiplied with a factor of 1.5 in case that reasoning models -- even if thinking is deactivated -- tend to reason during coding (e.g., Qwen or Gemma 4 models). We use a factor of 3.75 if thinking is activated, giving a max.\ context size of \num{15360}. During inference, we monitor whether any of these limits have been exceeded and re-run tests in case of significant impact on resulting performance measures.

Generated responses are post-processed based on the target evaluation criteria. 
For tasks requiring structural data retrieval, such as simulation component selection, the system requests an XML-like data structure, a format that state-of-the-art LLMs easily output. This markup text is validated and converted directly into native Python lists. In the event of an invalid or malformed response, errors are caught and logged inside the task structure, ensuring that downstream pipeline tasks remain unaffected.

\subsection{Code Execution}
\label{sec_codeExecution}
To rigorously evaluate LLM capabilities in synthesizing executable mechanical models, a robust code execution pipeline is required. When a model generates simulation scripts, the raw output undergoes structured post-processing prior to execution. First, the pure Python source code is programmatically isolated and extracted from the surrounding natural language text typically generated by the LLM. 

Following extraction, the pipeline enforces runtime safety checks on the raw code to prevent security hazards, focusing primarily on blocking illegal file system writes or unauthorized access to sensitive system packages. Once cleared, the code is dynamically augmented with special code lines and function calls which are designed to simplify subsequent evaluations, including the ability to identify different stages of errors, to redirect solver messages or outputs, and inject special settings for the generated finite element meshes. 
Crucially, the numerical solver initialization block is injected directly from the ground-truth file, guaranteeing identical numerical settings for the evaluated models. 
To safeguard the LLM pipeline against severe runtime errors and to apply hard execution timeouts, the generated script is executed within an independent, isolated Python subprocess. 
LLMs occasionally introduce logical anomalies, such as infinite loops, or synthesize erroneous code loops that instantiate a large number of bodies, causing the simulation to run for an unacceptable amount of time. 
Most importantly, it allows the framework to survive segmentation faults which sometimes occurred during meshing of the flexible parts with Netgen, triggered when an LLM passed geometrically illegal or self-intersecting boundary definitions to the meshing engine. 
After successful runs, all resulting metrics, status dictionaries, and mechanical states are then extracted and stored in the persistent, task-specific data layout for final statistical evaluation.


\section{Model Comparison and Verification} \label{sec_modelComparison}
\begin{figure}[tbh!]
    \centering
    \pgfsetlayers{background, behind, main}
    %
    %
    \begin{minipage}[t]{0.45\textwidth}
    \centering
    \textbf{a) levels~1~\&~2}\\[4pt]
    \begin{tikzpicture}[
        node distance=0.32cm,
        >=Stealth,
        block/.style={rectangle, draw=primary!30, fill=bglight,
            minimum width=3.4cm, minimum height=0.58cm, align=center,
            rounded corners=4pt, line width=0.7pt, font=\sffamily\scriptsize},
        startend/.style={rectangle, draw=primary, fill=primary, text=white,
            minimum width=3.4cm, minimum height=0.58cm, align=center,
            rounded corners=4pt, font=\sffamily\bfseries\scriptsize},
        failnode/.style={rectangle, draw=red!60, fill=red!8, text=red!70,
            minimum width=0.9cm, minimum height=0.45cm, align=center,
            rounded corners=3pt, font=\sffamily\scriptsize},
        line/.style={draw=primary!70, ->, line width=1.0pt},
        failline/.style={draw=red!55, ->, line width=0.7pt, dashed}
    ]
        \node[startend]          (S)  {rigid-body models};
        \node[block, below=of S]  (N1) {choose simulation components};
        \node[block, below=of N1] (N2) {finished assemble};
        \node[block, below=of N2] (N3) {topology match};
        \node[block, below=of N3] (N4) {graph content match};
        \node[block, below=of N4] (N5) {solver success};
        \node[block, below=of N5] (N6) {solution match};
        \node[startend, below=of N6] (E)  {overall success};

        \node[failnode, right=0.5cm of N2] (F2) {fail};
        \node[failnode, right=0.5cm of N5] (F5) {fail};
        \node[failnode, right=0.5cm of N6] (F6) {fail};

        \path[line] (S)  -- (N1);
        \path[line] (N1) -- (N2);
        \path[line] (N2) -- (N3);
        \path[line] (N3) -- (N4);
        \path[line] (N4) -- (N5);
        \path[line] (N5) -- (N6);
        \path[line] (N6) -- (E);

        \path[failline] (N2.east) -- (F2.west);
        \path[failline] (N5.east) -- (F5.west);
        \path[failline] (N6.east) -- (F6.west);
    \end{tikzpicture}
    \end{minipage}
    \hfill
    %
    %
    \begin{minipage}[t]{0.45\textwidth}
    \centering
    \textbf{(b) levels~3a~\&~3b}\\[4pt]
    \begin{tikzpicture}[
        node distance=0.32cm,
        >=Stealth,
        block/.style={rectangle, draw=primary!30, fill=bglight,
            minimum width=3.4cm, minimum height=0.58cm, align=center,
            rounded corners=4pt, line width=0.7pt, font=\sffamily\scriptsize},
        stagesep/.style={rectangle, draw=llmbeam!50, fill=llmbeam!10,
            minimum width=3.4cm, minimum height=0.58cm, align=center,
            rounded corners=4pt, line width=0.8pt,
            font=\sffamily\bfseries\scriptsize, text=llmbeam!80!black},
        startend/.style={rectangle, draw=primary, fill=primary, text=white,
            minimum width=3.4cm, minimum height=0.58cm, align=center,
            rounded corners=4pt, font=\sffamily\bfseries\scriptsize},
        failnode/.style={rectangle, draw=red!60, fill=red!8, text=red!70,
            minimum width=0.9cm, minimum height=0.45cm, align=center,
            rounded corners=3pt, font=\sffamily\scriptsize},
        line/.style={draw=primary!70, ->, line width=1.0pt},
        failline/.style={draw=red!55, ->, line width=0.7pt, dashed}
    ]
        \node[startend] (S)  {flexible-body models};
        \node[block, below=of S]  (P0) {choose part components};
        \node[block, below=of P0]  (P1) {interface name match};
        \node[block, below=of P1] (P2) {material match};
        \node[block, below=of P2] (P3) {mass match};
        \node[block, below=of P3] (P4) {mesh match};
        \node[block, below=of P4] (P5) {eigenfrequencies match};
        \node[startend, below=of P5] (P6) {part overall success};
        \begin{pgfonlayer}{behind}
	        \node[block, at=(P0), xshift=0.1cm, yshift=0.1cm] {choose part components};
	        \node[block, at=(P1), xshift=0.1cm, yshift=0.1cm] {interface name match};
	        \node[block, at=(P2), xshift=0.1cm, yshift=0.1cm] {material match};
	        \node[block, at=(P3), xshift=0.1cm, yshift=0.1cm] {mass match};
	        \node[block, at=(P4), xshift=0.1cm, yshift=0.1cm] {mesh match};
	        \node[block, at=(P5), xshift=0.1cm, yshift=0.1cm] {eigenfrequencies match};
	        \node[startend, at=(P6), xshift=0.1cm, yshift=0.1cm] {part overall success};
	    \end{pgfonlayer}
        \node[block, below=of P6]  (N1) {choose simulation components};
        \node[block, below=of N1] (N2) {finished assemble};
        \node[block, below=of N2] (N3) {topology match};
        \node[block, below=of N3] (N4) {graph content match};
        \node[block, below=of N4] (N5) {solver success};
        \node[block, below=of N5] (N6) {solution match};
        \node[startend, below=of N6] (E)  {overall success};
        \node[failnode, right=0.5cm of P6] (F1) {fail};
        \node[failnode, right=0.5cm of N2] (F2) {fail};
        \node[failnode, right=0.5cm of N5] (F5) {fail};
        \node[failnode, right=0.5cm of N6] (F6) {fail};
        \path[failline] (P6.east) -- (F1.west);
        \path[line] (S)  -- (P0);
        \path[line] (P0)  -- (P1);
        \path[line] (P1) -- (P2);
        \path[line] (P2) -- (P3);
        \path[line] (P3) -- (P4);
        \path[line] (P4) -- (P5);
        \path[line] (P5) -- (P6);
        \path[line] (P6) -- (N1);
        \path[line] (N1) -- (N2);
        \path[line] (N2) -- (N3);
        \path[line] (N3) -- (N4);
        \path[line] (N4) -- (N5);
        \path[line] (N5) -- (N6);
        \path[line] (N6) -- (E);
        \path[failline] (N2.east) -- (F2.west);
        \path[failline] (N5.east) -- (F5.west);
        \path[failline] (N6.east) -- (F6.west);
    \end{tikzpicture}
    \end{minipage}
    \caption{Sequential evaluation pipeline for level 1 and level 2~(a), as well as level 3~(b) tasks. Each step gates downstream evaluation; a failed check (dashed arrow, right) records the metric and terminates further evaluation for that task. For the flexible MBSs, special part metrics are evaluated for each part required for the MBS, illustrated using staggered arrangement of the boxes.}
    \label{fig:comparisonPipeline}
\end{figure}
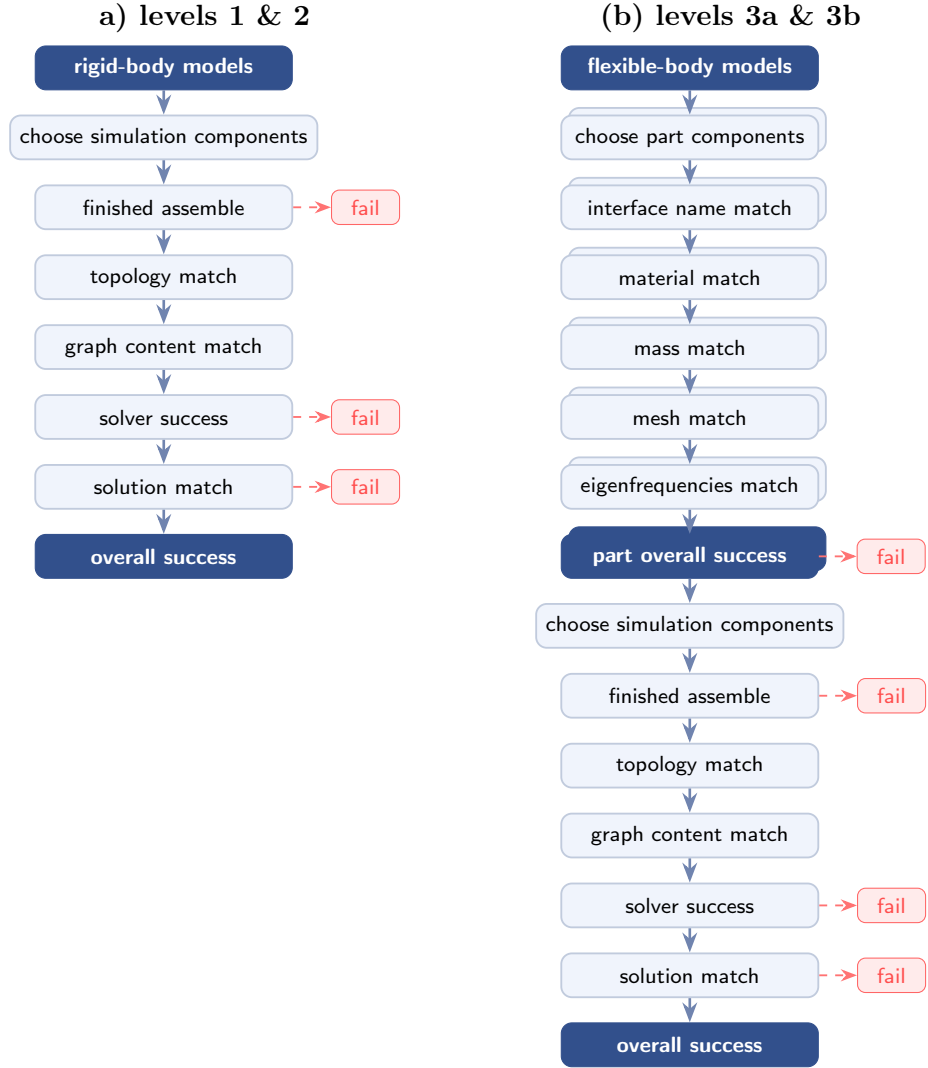

The evaluation framework in the present paper compares LLM-generated simulation scripts against expert-authored ground-truth scripts.
The evaluation uses a comprehensive set of metrics to quantify both the execution success and the physical accuracy of the generated models, see \fig{fig:comparisonPipeline}.
As we use the same LLM for all parts of a task, any error in the generation pipeline will be captured by exactly one metric.
In particular, we distinguish between the successful context retrieval for simulation model generation (choose simulation components), which is purely text-based, the successful execution of parts of the simulation model, the comparison of data structures in the simulation model, and the simulation results.

The following list gives an overview of different evaluation metrics:
\begin{itemize}
    \item \textbf{choose simulation components}: Success in selecting the correct simulation components from a predefined list of available components; if any component is missing (including components automatically added as dependencies), the task is counted as failed. If there are surplus components, the task is not counted as failed.
    \item \textbf{reached assemble, finished assemble}: Measures if the LLM-generated code reached and completed the system's \texttt{mbs.Assemble()} stage in the Exudyn engine or if an exception has been raised.
	Failure before assemble usually indicates a severe programming error -- often a hallucinated function argument, while an exception during assemble indicates an inconsistent system structure or invalid parameters; as both metrics agree very well, only \textbf{finished assemble} is shown.
    \item \textbf{solver success}: Evaluates the successful completion of the requested dynamic simulation over the given simulation period, in accordance with the ground truth model. Errors occur either during solver initialization, e.g., with kinematically impossible constraint conditions, or during the implicit time integration. Typical time integration errors are based on errors in the kinematics or in the system parameters.
    \item \textbf{topology match}: Uses graph isomorphism to verify if the LLM-generated connections (joints, interfaces) match the ground-truth topology, see \refSection{sec_graphBasedComparison} for details.
    \item \textbf{solution match}: Compares the numerical time-series output (body coordinates) of the LLM-generated simulation against the ground-truth simulation, see \refSection{sec_numericalSolutionComparison} for details.
    \item \textbf{graph content match}: Verifies the correctness of the attributes assigned to the graph nodes (e.g., mass, inertia, stiffness), see \refSection{sec_graphBasedComparison} for details.
    \item \textbf{overall success}: A final metric representing the successful completion of all required validation steps.
\end{itemize}

For level 3 tasks, a set of part-specific metrics is evaluated prior to the MBS metrics, covering geometrical and physical properties of the parts, as well as part interface names.
The part metrics are described in more detail in \refSection{sec_Part Metrics}. Be aware that the other metrics to evaluate level 3 tasks are exactly the same as for rigid-body models.

\subsection{Numerical Solution Comparison}\label{sec_numericalSolutionComparison}
The correctness of LLM-generated simulation codes is evaluated by a direct numerical time-series comparison of the body coordinates against the ground-truth data. The pipeline first verifies the structural dimensions, checking both the number of system coordinates and the number of simulated time steps. Wrong structural dimensions indicate major modeling differences, while a mismatch in time steps directly indicates a premature solver termination. 
For dimensionally matching simulations, the comparison uses the total coordinates $\mathbf{q}_{\mathrm{total}} = \mathbf{q}_{\mathrm{ref}} + \mathbf{q}_{\mathrm{disp}}$, since only the total solution, but not the reference state $\mathbf{q}_{\mathrm{ref}}$ or displacement $\mathbf{q}_{\mathrm{disp}}$ in isolation, determines the validity of the mechanical state. The discrepancy between generated and reference trajectories is quantified by the Frobenius norm of the full coordinate matrix across all time steps, with no normalization to number of time steps or coordinates applied -- as these dimensions do not change dramatically across the different tasks. Solutions are identified to be different, if this norm exceeds a tolerance of $10^{-5}$.

This relatively large tolerance is justified by the numerical nature of multibody dynamics. Minor round-off errors frequently arise when an LLM alters the order of adding spring-dampers or joints in the simulation code. Although algebraically identical, such reordering introduces tiny variations in the floating-point residuals and numerically computed Jacobians of some connectors. When these differences affect the convergence behavior or the exact number of Newton-Raphson iterations, a local numerical error occurs which is bounded by the solver's internal accuracies (relative tolerance of $10^{-8}$ and absolute tolerance of $10^{-10}$). Due to the non-linear error propagation inherent to dynamic time-integration, such a localized step error can magnify by several orders of magnitude over time, depending on the chaotic nature or stiffness of the underlying mechanical problem.

Finally, it must be noted that a direct time-series comparison is only valid if the coordinates map to identical physical entities. If the LLM generates bodies in an interchanged sequence compared to the sample code, the comparison fails and signals no solution match. To circumvent this without requiring complex graph-matching algorithms on the resulting matrices, the system prompt strictly instructs that the LLM adheres to a deterministic order of bodies, making the numerical solution directly comparable if the LLM follows instructions correctly.
Such sequence changes would be detectable, since the graph-based comparison is independent of the indexing of body numbers and thus receives a higher score for such cases, if the system is correctly parameterized. In our investigations, however, we did not observe a significant amount of such cases, see comparable success rates of numerical solution match and graph content match in \mytab{tab:mainResults}.

\subsection{Graph-Based Comparison}
\label{sec_graphBasedComparison}
To assess the topological correctness and parameter accuracy of LLM-generated configurations independently of their numerical solution or textual representation, a formal graph-based verification is implemented. 
Note that graph-representations of mechanical models are not unique, e.g., a mass-spring-damper system could be modeled equivalently by using one combined spring-damper or a separate spring and damper connector, giving different graphs.
Therefore, we generally expect to get the simplest representation from the LLM and we additionally compare the numerical results and monitor excessive performance differences for certain tasks.

Within the Exudyn multibody framework, simulation models are structurally defined via discrete computational elements, termed \textit{items}. These items comprise nodes, objects, markers, loads, and sensors, which collectively encapsulate all physical and kinematic properties of the mechanical system without relying on global environment variables.

By design, relationships between items follow a strict, limited set of connection rules that naturally define a directional graph topology:
\begin{itemize}
    \item \textbf{objects} (e.g., bodies, finite elements) map to underlying \textbf{nodes} to inherit degrees of freedom, and attach to \textbf{markers} to facilitate constraints or joints.
    \item \textbf{markers} reference specific \textbf{nodes} or \textbf{objects} to define localized operational points or coordinate systems.
    \item \textbf{loads} are applied exclusively by referencing predefined \textbf{markers}.
    \item \textbf{sensors} can monitor state variables across all item types, including recursive links to other \textbf{sensors}.
\end{itemize}

Using this framework, a unique mathematical graph is constructed from the MBS. The individual items serve as the graph nodes, while their internal references are translated into graph edges -- thus converting order-dependent indexing (e.g., an object pointing to a node index) into invariant structural graph edges, rendering the resulting topology fully independent of the arbitrary generation or index sequence chosen by the LLM. 

Furthermore, the full set of Exudyn item parameters -- except the above mentioned indices -- is embedded within the nodes via dictionary-based annotations. These annotations store all explicit physical attributes (e.g., mass, inertia tensors, spring stiffness, damping coefficients, and initial kinematic conditions) and parameters of joints and connectors. 
The instantiation, processing, and isomorphic verification of these structures are managed using the networkx\footnote{\url{https://github.com/networkx/networkx}} Python library, enabling separate validation of purely topological structures versus fully annotated graphs.

Certain specialized numerical parameters require explicit exceptions within this framework:
Python user-functions (they are only captured in numerical results), and numerically computed matrices, especially those arising from modally reduced flexible bodies, depending on the eigensolver's tolerances. Because flexible bodies are validated separately, these reduced structural matrices are omitted in the graph comparison.

The graph-based validation utilizes three levels, from basic counts to annotation-based comparison, proceeding only in case that the previous level was successful:
\begin{enumerate}
    \item \textbf{node and edge counts}: A preliminary, low-cost complexity check ensuring the total number of components and connections within the LLM-generated and the ground-truth graph.
    \item \textbf{topology match (graph isomorphism)}: A structural correctness check executing graph isomorphism algorithms. This layer evaluates whether the underlying network connectivity and categorical item types (e.g., distinguishing a revolute joint from a prismatic joint) match identically, disregarding parameter values.
    \item \textbf{annotation match}: An exact validation comparing the attributes of each node. If all annotations are identical, we know that the two multibody systems are identical.
\end{enumerate}

Due to randomization of some tasks, like long chains or meshes, the number of nodes may be excessively high, thus leading to an unacceptable duration of several stages in the graph comparison. For this reason, we add a maximum number of nodes for the comparison as well as a timeout of two minutes. We monitor any exceeding and timeout in the overall results and adjust model tasks accordingly to avoid such events.

To understand the nature of annotation differences, we cluster specific item types and parameter names (like body $\rightarrow$ physicsInertia), into specific physical and kinematic rubrics:
\begin{itemize}
    \item \textbf{kinematics / marker parameters}: Mainly errors in defining quantities related to kinematics by markers positions on bodies, leading to wrong kinematics.
    \item \textbf{rigid body configuration}: Errors in the configuration of rigid body references (we avoid to use initial values in the contextual information, thus reference configurations of rigid bodies also include initial displacements and rotations).
    \item \textbf{initial velocities}: Incorrect initialization of body velocities.
    \item \textbf{inertia and mass parameters}: Failures in assigning correct mass or inertia tensor values, usually due to wrong choice of body shape.
    \item \textbf{gravity and load parameters}: Errors in defining gravitational or external loading conditions to bodies.
    \item \textbf{connector / constraint parameters}: Mistakes in defining joints (like axis orientation or distance), or stiffness, damping, and further parameters in connectors.
    \item \textbf{other}: Miscellaneous errors not covered by the above categories.
\end{itemize}

For the evaluation, we count the number of occurrences of different kinds of errors, but no repeated occurrences within one task. This means that a chain of bodies, all of them with wrong mass, would only receive one mass-related error. However, there could be a connector parameter error in the same task. This allows us to evaluate the error statistics.

\subsection{Part Metrics}
\label{sec_Part Metrics}
For flexible machine parts, six dedicated metrics evaluate whether the meshed geometry, material assignment, and assigned interfaces are correct.
\begin{itemize}
	\item \textbf{choose part components}: Checks whether the correct components needed to generate the part are selected. Missing components result in a failed task, while surplus components are not penalized. Note that for the primitive geometry components ``box'' and ``cylinder'', the ``extrusion'' component is considered equally correct, since a box or cylinder can be created by extruding a rectangle or circle, respectively.
	\item \textbf{interface name match}: Verifies whether in the LLM generated part both the number and names of selected interfaces are correct.
	\item \textbf{material match}:  Evaluates whether the correct Young's modulus, Poisson's ratio, and density have been assigned; all three must match exactly.
	\item \textbf{mass match}: Verifies whether the mass of the LLM-generated part lies within 1\% of the mass of the ground truth part, providing a coarse check of overall geometry without penalizing minor deviations such as incorrect fillet radii.
	\item \textbf{mesh match}: Computes and validates the per-vertex distance between the LLM-generated and ground truth meshes via PyMeshLab\footnote{\url{https://github.com/cnr-isti-vclab/PyMeshLab}}, normalized by the ground truth volume, with a 1\% tolerance.
	\item \textbf{eigenfrequencies match}: 
	If fewer than six eigenfrequencies are requested by the LLM, this is silently corrected before execution (in order to make solutions comparable). The first three eigenfrequencies are then compared with ground truth values using a relative vector-norm difference with a 5\% tolerance. This metric jointly captures geometric correctness through the resulting stiffness and mass matrices and correct interface placement, since the chosen interfaces influence the computed modes.
\end{itemize}
As described in \refSection{sec_Flexible Multibody Systems Assembly}, only if all part metrics pass, the pipeline proceeds to the flexible MBS assembly phase.

\section{Results and Statistical Evaluation}
\label{sec:results}

Experiments were executed locally on a Windows~11 workstation (64\,GB RAM) equipped with an NVIDIA GeForce RTX~5090 GPU (32\,GB VRAM) and an Ubuntu 24.04 server equipped with two NVIDIA H100 GPUs (80\,GB VRAM each).
The main Python packages were Ollama~0.24.0 and Exudyn~1.10.160.

\subsection{Simple and Medium Rigid-Body MBS Models}
Evaluation was first carried out on the combined set of 50 task models spanning difficulty levels~1 and~2, which allow a cleaner evaluation, since in level 3 the success of the system assembly is strongly conditioned on the preceding part generation.

Each LLM was tested on 20 independent random parameter variations per task model, yielding 1000 tasks per LLM run.
In addition to the main benchmark, dedicated runs explored the effect of temperature, reasoning, and prompt template variations; these are discussed in Sections~\ref{sec_temperature_effects}--\ref{sec_prompt_effects}.

\subsubsection{Overall Performance}
A total of 32 open-weight LLMs and two proprietary LLMs were evaluated, spanning release dates from late 2023 to mid-2026 and covering a wide range of open-weight model sizes (less than 4\,B up to 122\,B effective parameters at 4-bit quantization).
\mytab{tab:mainResults} reports the success rates at each stage of the evaluation pipeline for all tasks, LLMs sorted by overall success rate, using Ollama LLM names throughout the paper.
\begin{table}[htb]
  \centering
  \small
  \footnotesize
  \caption{Overall evaluation results for level 1 and 2 tasks with
         1000 tests for 10 best performing LLMs (full list in \mytab{tab:modelComparisonAll}),
				 using zero temperature and no reasoning, except gpt-oss\textsuperscript{$\dagger$}, and proprietary LLMs which were running with default API settings; rates are given in \%. 
         $\dagger$gpt-oss LLMs using minimum reasoning effort (setting: ``low'').}
  \label{tab:mainResults}
\begin{tabular}{l@{\hspace{-5pt}}rrrrrr}
\toprule
 & \multicolumn{6}{c}{\textbf{success rate (\%)}} \\
\cmidrule(lr){2-7}
\textbf{LLM} & \textbf{comp.} & \textbf{solver} & \textbf{topol.} & \textbf{numsol} & \textbf{graph} & \textbf{overall} \\
\midrule
Claude-Opus-4.8 & 99.4 & 99.7 & 99.2 & 96.1 & 93.2 & 91.4 \\
GPT-5.2 & 98.6 & 98.9 & 98.2 & 91.5 & 90.3 & 86.0 \\
\midrule
qwen3.6:27b & 95.8 & 94.0 & 91.0 & 83.3 & 83.3 & 82.1 \\
gemma4:31b & 97.0 & 95.9 & 94.3 & 84.3 & 84.0 & 81.1 \\
qwen3.5:27b & 96.6 & 96.0 & 93.9 & 83.1 & 83.5 & 81.1 \\
gpt-oss:120b\textsuperscript{$\dagger$} & 97.6 & 89.5 & 86.9 & 77.9 & 75.2 & 73.3 \\
qwen3.5:35b & 95.5 & 84.6 & 85.3 & 73.2 & 71.7 & 70.3 \\
gpt-oss:20b\textsuperscript{$\dagger$} & 97.8 & 87.8 & 85.3 & 72.0 & 70.0 & 67.6 \\
gemma4:26b & 99.9 & 84.7 & 84.5 & 70.2 & 66.6 & 66.0 \\
qwen3.5:122b & 86.6 & 88.3 & 87.6 & 76.4 & 74.1 & 62.5 \\
\midrule
mean (proprietary) & 99.0 & 99.3 & 98.7 & 93.8 & 91.8 & 88.7 \\
mean (top 8 open-weight) & 95.9 & 90.1 & 88.6 & 77.5 & 76.0 & 73.0 \\
mean (all 32 open-weight) & 71.9 & 68.3 & 57.6 & 44.1 & 41.0 & 37.1 \\
\bottomrule
\end{tabular}

{\footnotesize comp. = choose simulation components; solver = solver success; topol. = topology match;
numsol = numerical solution match; graph = graph content match}
\end{table}

The best open-weight LLM, qwen3.6:27b, achieves an overall success rate of 82.1\%, with gemma4:31b and qwen3.5:27b very close, within the margin of statistical noise.
Following the full \mytab{tab:modelComparisonAll}, seven LLMs score below 10\%, because early LLMs are not able to follow instructions in component selection (qwen:32b).

The performance of the top eight open-weight models achieves a remarkable 73\%, noting a significant performance drop between topology match (88.6\%) and the remaining success rates, like numerical solution (77.5\%) and graph content match (76.0\%).
This performance drop, together with detailed analyses of failure cases, indicates that LLMs are able to capture main mechanical approaches and concepts, but often fail in fine-grained selection of mechanical parameters.

The detailed performance values in \mytab{tab:modelComparisonAll} highlight different qualities of LLMs, even within the same model families, such as gemma4:26b which outperforms all other LLMs (including gemma4:31b) in simulation component selection.
Similarly, ministral-3 shows a high component selection rate while its overall performance is below top-10, attributed to its low mechanical and coding capabilities.

Beyond accuracy, the models differ in runtime by more than an order of magnitude, compare the last column in \mytab{tab:thinking_effects} without reasoning, which are the runtimes equivalent to \mytab{tab:mainResults}. 
On a single NVIDIA RTX 5090, gpt-oss:20b and gemma4:26b complete the generation in 18 and 33 minutes, respectively, whereas the higher-scoring qwen3.5:27b and qwen3.6:27b require 167 and 235 minutes. With appropriate Ollama settings, the gemma and gpt-oss models support batched inference, while as of this writing, Ollama does not implement batched inference for qwen3.5/3.6, which therefore do inference sequentially.
Additionally, gpt-oss:20b and gemma4:26b lead to very short runtimes due to mechanisms such as sliding-window attention and mixture-of-experts, which improves caching. We like to mention the high performance, as it helped us to develop the framework with very short test runs for the latter two models.

\mytab{tab:errorCategories} includes the detailed graph content errors for five selected models (gemma4:26b, gemma4:31b, gpt-oss:20b, qwen3.5:27b and qwen3.6:27b), with several rubrics for the types of Exudyn parameters which are incorrect.
As the proportions show for the basic mode (no reasoning), error rubrics are all in a similar range except for loads.
Comparing results with and without reasoning, the average count reveals that kinematics, constraint, rigid body configuration and inertia-related errors significantly drop, generally making the model set up more correct.
The remaining initial velocity errors, mainly appearing for rolling wheels, are resolved by more capable proprietary LLMs (for Claude-Opus-4.8, it gives 3 velocity errors).

\subsubsection{LLM Performance versus Release Date}\label{sec_performance_release_date}
\begin{figure}[htb]
    \centering
    \begin{subfigure}[t]{0.495\textwidth}
        \centering
        \includegraphics[width=\textwidth]{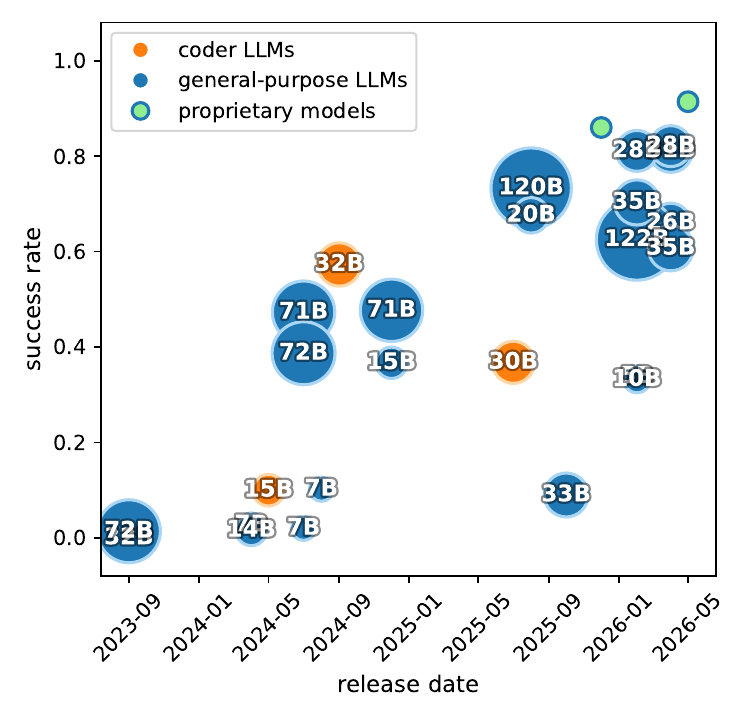}
        \caption{Overall success rate}
        \label{fig:performanceScatterPlot50models_a}
    \end{subfigure}
    \hfill
    \begin{subfigure}[t]{0.495\textwidth}
        \centering
        \includegraphics[width=\textwidth]{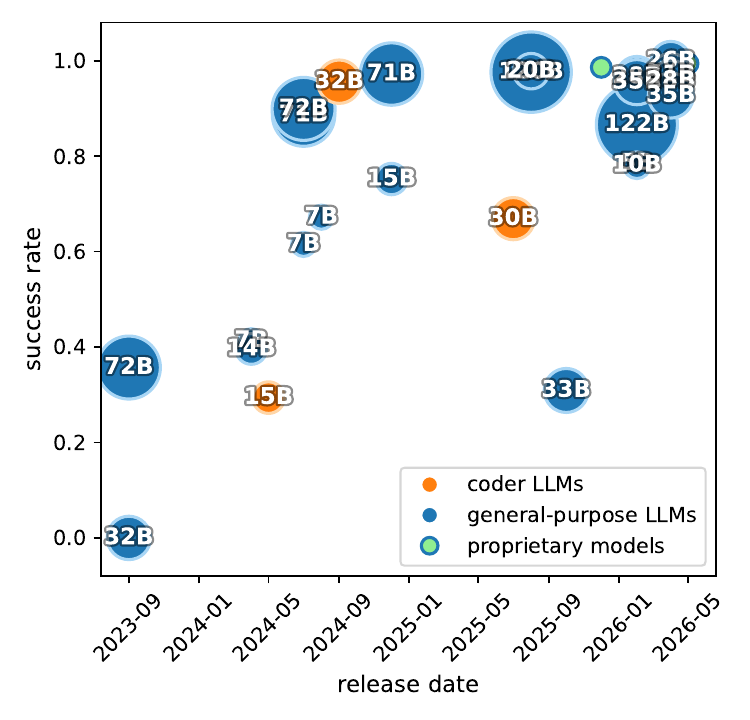}
        \caption{Component-selection success rate}
        \label{fig:performanceScatterPlot50models_b}
    \end{subfigure}
    \caption{Success rates of the evaluated LLMs versus release date. All models were run at zero temperature, except proprietary models, which used their default API settings. Circle areas are proportional to the number of model parameters, proprietary models with publicly unknown parameter counts are shown as green circles. Models released in 2026 consistently outperform earlier generations.}
    \label{fig:performanceScatterPlot50models}
\end{figure}

\fig{fig:performanceScatterPlot50models}a shows the temporal trend of the overall success rate, while \fig{fig:performanceScatterPlot50models}b shows the corresponding trend for the component selection success, clearly demonstrating the ongoing increase of LLMs' capabilities for the given mechanical tasks.
Without clear evidence, we mention instruction following and coding, which was also identified to be related to the success of such tasks in our earlier work \cite{Moeltner2026_CreationEvaluationSelfValidation}, potentially in combination with improved training and finetuning on physics and math tasks.
Specifically, we see that coding models can outperform much larger models of their time, compare orange circles in \fig{fig:performanceScatterPlot50models}.
It can also be observed that for simulation component choice, LLMs from mid-2024 can reach very high success rates.
Worth mentioning, selected proprietary LLMs outperform their counterparts, however, the clear trend of improvements of open-weight LLMs indicates a six to twelve months ``delay'' in equal performance.

\subsubsection{Temperature Effects}\label{sec_temperature_effects}
To assess the sensitivity of mechanical engineering capabilities to the sampling temperature, six LLMs from \mytab{tab:mainResults} -- five of the top-performing models together with ministral-3:14b, included to probe whether temperature sensitivity differs for a lower-ranked model -- were evaluated on the level 1 and level 2 task sets at eleven equidistant temperature values from 0 to 1, plus an additional value of $0.05$ to capture effects at very low temperatures. 

The two Gemma 4 models are essentially insensitive to temperature, which is attributed also to the design of these newer models with recommendations to use temperature values even larger than 1 for tasks requiring more ``creativity''.
All other LLMs, in contrast, show a clear downward trend as temperature increases in case that no reasoning is used.
As commonly known for reasoning LLMs, a larger temperature leads to an increased performance, see the special case of gpt-oss:20b (think) results. This is the reason why we used zero temperature for non-reasoning but default temperature (usually around 1) for reasoning, e.g. in \mytab{tab:thinking_effects}.
ministral-3 shows the strongest degradation of the success rate, dropping from 54\% at temperature~0 to 42\% at temperature~1 (a loss of 12 percentage points). 
Overall, low temperatures are preferable, while robustness to temperature itself shall be regarded as a desirable property of a mechanically aware LLM.

\begin{figure}[htb]
    \centering
    \includegraphics[width=0.85\textwidth]{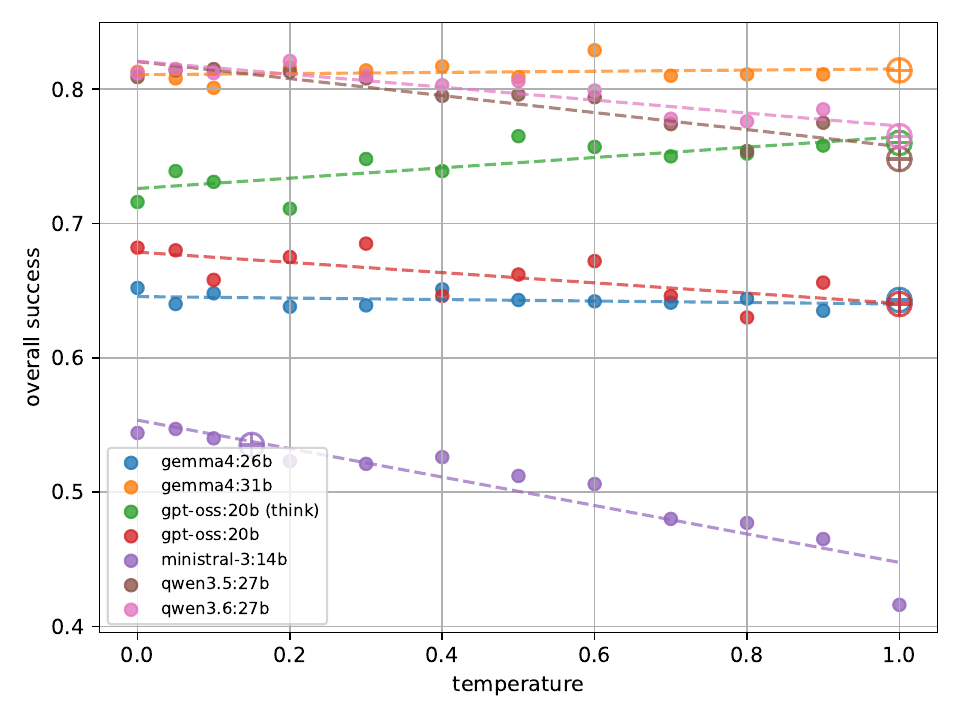}
    \caption{Overall success rate versus sampling temperature for six LLMs, evaluated on the full level 1 and 2 task sets; gpt-oss:20b is shown with ``low'' and with ``medium'' (think) reasoning; circled plus signs indicate results with default temperatures; dashed lines show linear trends per model.}
    \label{fig:temperatureCorrelation}
\end{figure}

\subsubsection{Thinking Effects}
More recently, LLMs have been equipped with reasoning, often denoted as thinking.
Typical LLMs like gpt-oss, Gemma 4 or Qwen 3.5 models include flags in their inference API which allow to activate reasoning, or to set the level of thinking (gpt-oss).

In common benchmarks \cite{guan2026cad, Zeng2025LEGOpuzzles}, LLMs achieve higher scores with activated reasoning.
However, the amount of response tokens significantly increases due to thinking, which can be orders of magnitude larger than the main response content -- in particular for very short responses like answers with yes or no.
The significantly larger number of tokens also impacts the response time, first because the response time is almost proportional to the number of tokens, and second because of a longer context size which again slows down the inference.

In the present investigation, we highlight the impact of thinking, as shown in \mytab{tab:thinking_effects}.
We observe that the choose simulation components score is positively affected by thinking, however, with drastic impact on the number of tokens from several dozens without reasoning to a few thousands with reasoning.
The overall success may be significantly affected by thinking, boosting the scores of gemma4:31b, gpt-oss:20b and gemma4:26b by 5\% to 9\%. 
As shown for qwen3.5/3.6, thinking can also degrade the performance for such tasks.
Looking at the error categories in \mytab{tab:errorCategories}, we observe that thinking reduces errors related to kinematics, reference conditions and inertia by factors of 3 to 5, indicating that thinking improves basic multibody reasoning.

However, we observe that the inference time can increase by a factor greater than 10 due to reasoning, such as for gemma4:26b and qwen3.6:27b, and the total number of response tokens is 21 times higher for qwen3.5:27b.
Therefore, we performed most overall tests without thinking, as inference time would not have allowed to complete results such as for temperature variations of qwen3.6:27b in acceptable time, while the average performance increases only moderately.

\begin{table}[tb]
  \centering
  \caption{Effects of reasoning on total number of tokens (response and
    reasoning), LLM inference time, and success rates for choose simulation components
    as well as overall success. Note that gpt-oss reasoning level was set to ``low''
    for non-reasoning and to ``medium'' for reasoning.}
  \label{tab:thinking_effects}
  \resizebox{\textwidth}{!}{
\begin{tabular}{l cc cc cc cc}
\toprule
 & \multicolumn{2}{c}{\textbf{choose comp.}} & \multicolumn{2}{c}{\textbf{overall success}} & \multicolumn{2}{c}{\textbf{total tokens (k)}} & \multicolumn{2}{c}{\textbf{LLM dur. (min.)}} \\
\cmidrule(lr){2-3}\cmidrule(lr){4-5}\cmidrule(lr){6-7}\cmidrule(lr){8-9}
\textbf{LLM} & thk. & no-thk. & thk. & no-thk. & thk. & no-thk. & thk. & no-thk. \\
\midrule
gemma4:26b & 99.8\% & 99.9\% & 75.0\% & 66.0\% & 6245 & 622 & 335 & 33 \\
gemma4:31b & 99.6\% & 97.0\% & 86.0\% & 81.1\% & 3267 & 521 & 670 & 109 \\
gpt-oss:20b & 98.5\% & 97.8\% & 76.0\% & 67.6\% & 2609 & 750 & 59 & 18 \\
qwen3.5:27b & 99.2\% & 96.6\% & 76.7\% & 81.1\% & 9659 & 454 & 2979 & 167 \\
qwen3.6:27b & 100.0\% & 95.8\% & 77.9\% & 82.1\% & 7213 & 664 & 2400 & 235 \\
\bottomrule
\end{tabular}
}
\end{table}

\begin{table}[htb]
  \centering
  \small
  \caption{Distribution of graph-content errors across categories, given as the
    proportion (\%) of all errors and as the average error count per model
    (mean over the five models shown in \mytab{tab:thinking_effects}),
    comparing reasoning (``thk.'') and non-reasoning (``no-thk.'') runs.}
  \label{tab:errorCategories}
\begin{tabular}{l cc cc}
\toprule
 & \multicolumn{2}{c}{\textbf{proportion (\%)}} & \multicolumn{2}{c}{\textbf{average count}} \\
\cmidrule(lr){2-3}\cmidrule(lr){4-5}
\textbf{error category} & thk. & no-thk. & thk. & no-thk. \\
\midrule
kinematics / marker parameters & 10.7 & 19.4 & 13.2 & 41.6 \\
rigid body configuration & 3.7 & 10.5 & 4.6 & 22.6 \\
initial velocities & 19.2 & 14.0 & 23.6 & 30.0 \\
inertia and mass parameters & 7.8 & 18.9 & 9.6 & 40.6 \\
gravity and load parameters & 1.0 & 0.7 & 1.2 & 1.6 \\
connector / constraint parameters & 23.1 & 18.8 & 28.4 & 40.2 \\
other & 34.4 & 17.6 & 42.2 & 37.8 \\
\bottomrule
\end{tabular}

\end{table}

\subsubsection{Prompt Design Effects}\label{sec_prompt_effects}

\begin{figure}[tb]
    \centering
    \includegraphics[width=0.8\linewidth]{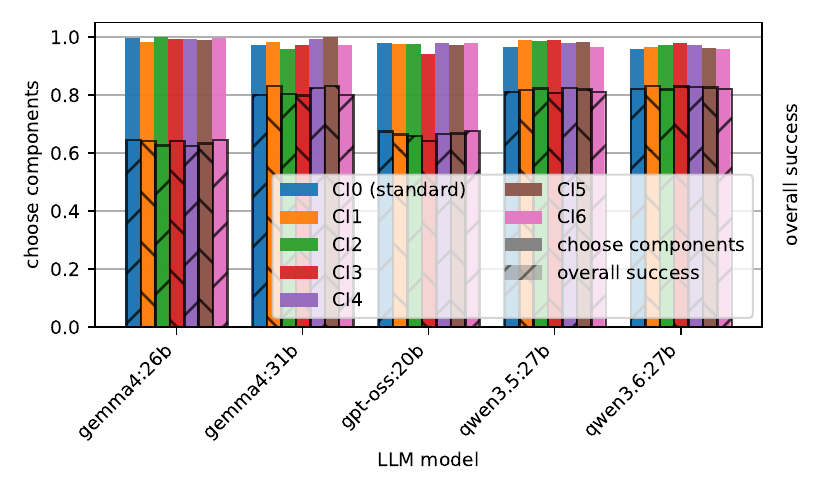}
    \caption{Effect of choose-component prompt variation (CI1--CI6) on component-selection rate (solid bars) and overall success rate (hatched bars) per LLM model.}
    \label{fig:pvci}
\end{figure}

\begin{figure}[htb]
    \centering
    \includegraphics[width=0.8\linewidth]{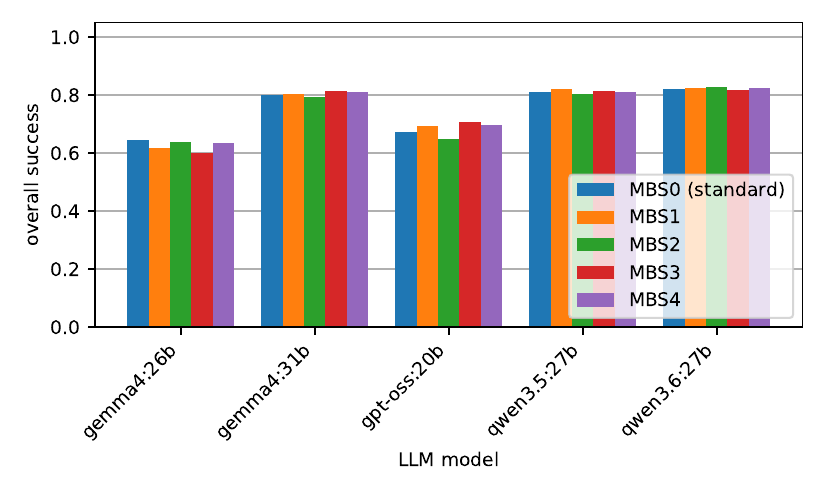}
    \caption{Effect of MBS code-generation prompt variation (MBS1--MBS4) on overall success rate per LLM model.}
    \label{fig:pvmbs}
\end{figure}

To assess the sensitivity to prompt formulation, two ablation studies were conducted: varying the prompt template for the component-selection stage (CI1--CI6) and for the code-generation stage (MBS1--MBS4), while keeping all other settings fixed.
The variations are either based on specific wordings 
The templates vary primarily in restrictiveness and structure, ranging from minimal guidance to explicit instructions, while keeping the task and output schema constant.
\fig{fig:pvci} shows the component-selection rates (solid bars) and overall success rates (hatched bars) for six alternative component-selection prompts. 
\fig{fig:pvmbs} shows the overall success rates across four alternative MBS code-generation prompts.
Both, the component-selection rates as well as the overall success rates show moderate variations across all variations, indicating that our LLM-pipelines are insensitive to prompt formulations.

\begin{table}[tb]
	\centering
	\small
	\caption{Overall evaluation results for level 3 tasks; each LLM is tested for its ability to generate 440 flexible parts and, if possible, 240 flexible MBSs. The part and MBS metrics are visually separated using a vertical line. For the open-weight models, the temperature is zeroed and reasoning is turned off. $\dagger$gpt-oss LLMs using minimum reasoning effort (setting: ``low'').}
	\label{tab:flexuAIOverall}
	\resizebox{\textwidth}{!}{
\begin{tabular}{l ccc|c|ccc}
\toprule
 & \multicolumn{7}{c}{\textbf{success rate (\%)}} \\
\cmidrule(lr){2-8}
\textbf{LLM} & part comp. & mesh & parts overall & MBS started & comp. & graph & overall \\
\midrule
Claude-Opus-4.8 & 100.0 & 83.0 & 82.7 & 70.8 & 70.8 & 66.2 & 65.4 \\
GPT-5.2 & 100.0 & 80.5 & 80.2 & 66.2 & 65.8 & 60.0 & 59.6 \\
\midrule
gemma4:31b & 100.0 & 77.5 & 76.8 & 62.1 & 62.1 & 52.1 & 52.1 \\
qwen3.5:27b & 95.2 & 60.5 & 55.9 & 42.5 & 42.5 & 32.5 & 32.5 \\
qwen3.6:27b & 97.3 & 68.9 & 64.1 & 43.8 & 41.2 & 30.4 & 30.0 \\
gpt-oss:20b\textsuperscript{$\dagger$} & 98.6 & 43.4 & 38.0 & 22.5 & 22.5 & 15.8 & 15.4 \\
ministral-3:14b & 100.0 & 47.5 & 39.6 & 24.2 & 24.2 & 12.5 & 12.5 \\
gemma4:26b & 95.5 & 39.8 & 30.0 & 10.4 & 8.8 & 5.0 & 5.0 \\
\midrule
mean (proprietary) & 100.0 & 81.7 & 81.5 & 68.5 & 68.3 & 63.1 & 62.5 \\
mean (open-weight, n=6) & 97.8 & 56.2 & 50.7 & 34.2 & 33.5 & 24.7 & 24.6 \\
\bottomrule
\end{tabular}
}
	{\footnotesize part comp. = choose part components; mesh = mesh match; MBS started = rate how many MBS assembly tasks are started}
\end{table}

\subsection{Flexible MBS Models}
As with the rigid-body MBSs, the results shown here were obtained by testing the ability of the investigated LLMs to create each of the 12 flexible MBSs 20 times with different, randomized parameters. In total, 240 flexible MBSs were generated. Since not every flexible MBS requires only one flexible part, some require two, as shown in \fig{fig:flexuAIModels2Part}, or even three, the 240 flexible MBSs require a total of 440 flexible parts.

The part metrics are always computed with respect to these 440 parts: if an LLM, for example, achieves 50\% on ``parts overall'', it has successfully generated 220 parts without any detected error. The MBS metrics, in turn, are always computed with respect to the 240 tested flexible MBSs. If at least one part requested by a flexible MBS is not evaluated as overall correct, the flexible MBS assembly phase is not started, see ``failed'' marks in \fig{fig:comparisonPipeline}.

\subsubsection{Overall Performance}
\mytab{tab:flexuAIOverall} shows the results for the six tested open-weight LLMs and the two proprietary LLMs, ordered by the overall correctness of the generated flexible MBS.

Both proprietary models and the best open-weight model, gemma4:31b, selected the needed components for all 440 parts without a single miss. Also ministral-3:14b, which already performed well at component selection for the level 1 and level 2 tasks, achieves full success here. On average, the proprietary models generated parts with an overall correctness of around 82\%, compared to around 51\% for the open-weight models. Reaching almost 77\% overall correctness of the generated parts, gemma4:31b performs well above the open-weight average.

Not all flexible MBS assemblies could be continued after the part generation phase, since the required parts were not always generated correctly. The amount of started assembly tasks is shown in the table. Remarkably, only around 10\% of the flexible MBS could be started when using gemma4:26b because of errors in most of the generated parts. As the results show, Claude-Opus-4.8, gemma4:31b, qwen3.5:27b, gpt-oss:20b, and ministral-3:14b then selected all of the required simulation components for the flexible MBS assemblies that were started. The drop in success rate within the flexible MBS assembly stage is small, being less than 10 percentage points for the average open-weight LLM. This is considerably smaller than the drop observed for level 1 and level 2 tasks, where it exceeded 20\% for the best eight open-weight models between the ``choose simulation components'' and ``overall success'' stages. This is because the assemblies themselves are easier for flexible MBSs than for level 1 and level 2 tasks.

Overall, Claude-Opus-4.8 achieved the highest success rate at around 65\% in setting up flexible MBS simulations from textual descriptions, with GPT-5.2 following more than 5 percentage points behind. Among the open-weight models, gemma4:31b achieved the highest success rate, whereas gemma4:26b achieved only 5\% overall, falling well below the open-weight average of approximately 25\%. This suggests that, at least for this specific LLM, parameter count plays a particularly critical role for such complex multibody dynamics tasks.

\begin{table}[htbp]
	\centering
	\caption{Statistics of the LLMs used to solve level 3 tasks, ordered by the overall success of the part generation phase. For the open-weight models, the temperature is set to zero and reasoning is turned off. Information marked with ``-'' is not available for the proprietary models. $\dagger$gpt-oss LLMs using minimum reasoning effort (setting: ``low'')}
	\label{tab:flexuAIInfo}
	\resizebox{\textwidth}{!}{
\begin{tabular}{lcccccc}
\toprule
 & \multicolumn{6}{c}{\textbf{LLM statistics}} \\
\cmidrule(lr){2-7}
\textbf{LLM} & LLM dur. & parts eval. dur. & MBS eval. dur. & gen. tokens (k) & ctx size used (P95) & batch size \\
\midrule
Claude-Opus-4.8 & 307.6 & 35.4 & 3.3 & 1562 & - & - \\
GPT-5.2 & 256.3 & 36.3 & 2.9 & 1272 & - & - \\
\midrule
gemma4:31b & 133.3 & 30.4 & 1.6 & 457 & 4734 & 10 \\
qwen3.5:27b & 174.2 & 28.0 & 1.1 & 566 & 2487 & 5 \\
qwen3.6:27b & 283.9 & 26.2 & 1.2 & 950 & 5568 & 5 \\
gpt-oss:20b\textsuperscript{$\dagger$} & 57.7 & 28.7 & 5.9 & 385 & 2160 & 5 \\
ministral-3:14b & 49.6 & 29.1 & 0.7 & 392 & 3004 & 4 \\
gemma4:26b & 51.4 & 27.7 & 0.2 & 489 & 1682 & 5 \\
\midrule
mean (proprietary) & 282.0 & 35.8 & 3.1 & 1417 & - & - \\
mean (open-weight, n=6) & 125.0 & 28.4 & 1.8 & 539 & 3272 & 5 \\
\bottomrule
\end{tabular}
}
\end{table}

\subsubsection{Evaluation and Inference Time}
Using the Windows 11 workstation with an RTX 5090 GPU, LLM duration and evaluation time were tracked and are reported in \mytab{tab:flexuAIInfo}, together with the total amount of the generated tokens and the 95 percentile (P95) of the used context window size. With the context limit set to 6144 tokens for level 3a and level 3b tasks, the observed context usages show the context size limit is not a constraint for the majority of workloads.

On average, the LLM duration required was around 2 hours for the open-weight models. Evaluating the correctness of the parts using the metrics described in \refSection{sec_Part Metrics} took approximately half an hour, while checking the assemblies according to the metrics described in \refSection{sec_modelComparison} took about two minutes. Consequently, for an open-weight model we expect a typical runtime of 2 to 3 hours; for proprietary models, longer -- though these values are not directly comparable, as API inference times depend on network latency, server-side load, and rate limits rather than solely on model speed. 

\begin{figure}[tbh]
	\centering
	\includegraphics[width=1.0\textwidth]{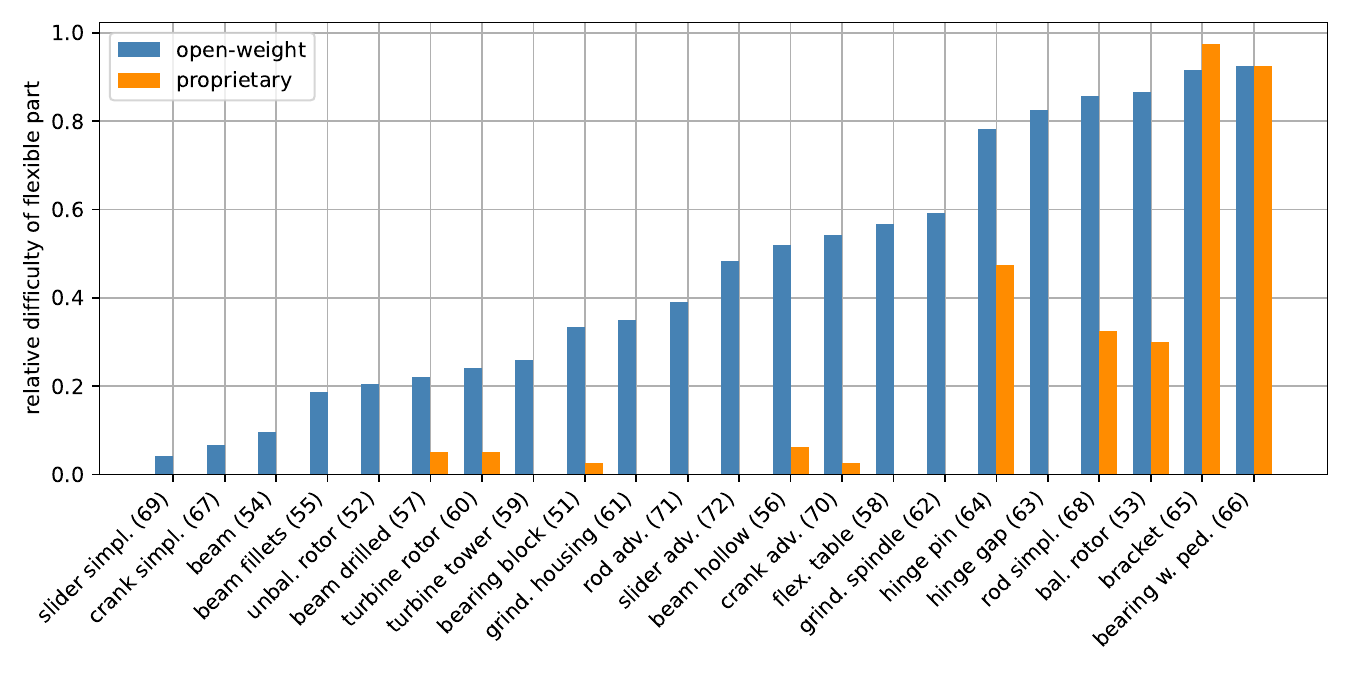}
	\caption{Difficulty of the generated parts for the tested LLMs shown in \mytab{tab:flexuAIOverall}. Open-weight and proprietary models are shown separately.}
	\label{fig:flexuAIDifficulty}
\end{figure}

\subsubsection{Difficulty of Tasks}
This section evaluates the difficulty of the tasks. While the level 1 and level 2 tasks focus on the MBSs, here the focus is on part generation. \fig{fig:flexuAIDifficulty} shows the difficulty of the parts for the LLMs, with parts ordered according to the difficulty they pose for the open-weight models. Difficulty is defined as $1-(\mbox{mean overall success rate})$ achieved by the open-weight and proprietary models for a given part.
A difficulty of 0, for example, would mean that all LLMs generated every required instance of the part correctly.

In general, there are 12 parts that are made correctly every time by the proprietary models, whereas no part is generated correctly every time by the open-weight models. However, there is also no part that is never generated correctly by either class of models. This indicates that all part-generation tasks are, in principle, solvable by both classes of LLMs, but some specific LLMs never manage to solve a given part, for example, Claude-Opus-4.8 with the ``bearing block with pedestal'' part, and gemma4:31b with the ``mounting bracket'' part. However, the parts are also relatively complex, so difficulties close to 1 are still observed.

Across the part metrics, we observed the largest drops in success rate after component selection, specifically at the three geometry-related metrics: mass match, mesh match, and eigenfrequencies match. This indicates that correct geometric assembly remains the primary bottleneck. Notably, both the proprietary and open-weight models struggle with the ``mounting bracket part'' and the ``bearing block with pedestal part''. Both parts require a chamfer on specific edges, and even otherwise high-scoring models such as gemma4:31b, Claude-Opus-4.8, and GPT-5.2 have difficulty assigning these edges correctly. 
%

Another issue worth reporting concerns the setting of reference positions when assembling the ``flexible table with load'' MBS, which has an average overall success of only 32.5\%, while the ``flexible table part'' alone has an overall success of more than 50\%. The ``flexible table part'' is created such that the center of mass of the table's plate lies at the origin. In the assembly, the LLMs must place the flexible part so that the bottom of the table's feet touch the ground plane, correctly requiring an offset equal to the feet's height plus half the plate's height. However, many tested LLMs shift the table by the feet's height alone. This underlines: the MecEng benchmark tests the LLMs' mechanical awareness.



The lower success rates of level 3 tasks, relative to level 1 and level 2, stem from the sequential nature of the task setup: finite element meshes must be correctly generated for all constituent parts, and any single error causes the entire task to fail. The LLM must execute every step correctly, since even a minor error in the part-generation phase compromises the entire assembly. This mirrors real-world mechanical engineering and multibody dynamics practice, where an engineer must likewise execute every step correctly to arrive at a valid result. This sequential dependency underscores why the MecEng benchmark is particularly well suited to benchmarking the mechanical awareness of LLMs, while also revealing clear limitations of state-of-the-art models. This finding will be revisited in the closing discussion.

\section{Discussion, Limitations and Outlook}
\label{sec:limitations}
The present paper represents a big step from previous research on simple multibody systems to complex mechanism and flexible body models.
The overall results not only show the enormous potential of LLMs for mechanical tasks, but also the severity of LLM errors in creating engineering simulation models. While some LLM-generated simulation models are inconsistent or do not follow the coding instructions, leading to coding-related errors reported to the user (9.9\% of tasks averaged over the top 8 open-weight LLMs have no solver success, see \mytab{tab:mainResults}), there is a significant amount of cases in which the LLMs produce executable simulation models, however, with hard-to-detect physical or kinematic errors (17.1\% of tasks averaged over the top 8 open-weight LLMs, difference of solver and overall success in \mytab{tab:mainResults}).
As the coding-related error rates do not dominate and because we can clearly distinguish from the mechanics-related errors, we conclude that the reported success rates of the MecEng benchmark provide an excellent indicator for {\it Mechanical Engineering Awareness}.

While we extended the previous purely numerical evaluation to component selection, graph comparison and assembly- and solver-error metrics, the comparison still remains challenging for certain classes of problems. In particular, comparison of user functions requires a testing of such user functions (e.g. at least for a set of randomized parameters).
Evaluating a large number of tests reveals detailed statistics with excellent reproducibility, however, the tests require fully automated runs. Failures due to (nearly) infinite loops, bad solver convergence, exponentially growing effort for graph isomorphism checks and segmentation faults of the external mesher have to be treated carefully.
In the current implementation, we monitor all such events within a final results overview and resolved errors accordingly or revised the model parametrization such that they do not produce timeouts.

The test runs produce an enormous amount of data: for performance plots in \fig{fig:performanceScatterPlot50models} the collected data comprises \num{94\,565} files and 2.9\,GB of text, which made human evaluation impossible.
Irrespective of the big-data problem, we at least manually monitored those tasks which were not completed by the best (proprietary) LLMs; this required several iterations regarding clear definition of tasks and correct ground-truth implementation.
Herein, we must admit that the best proprietary LLM uncovered roughly as many human errors in task definitions and ground-truth implementations as itself makes when it generates level 1 and 2 simulation models.

Future agentic frameworks, e.g., motivated by our earlier self-validation approach \cite{Moeltner2026_CreationEvaluationSelfValidation}, may remove simulation model errors by iterative model improvements and they may detect modeling errors by analyzing simulation outputs or image-based initial or current system configurations, using snapshots as shown in \fig{fig:resultsExamples}. Another approach could use the comparison of independently generated simulation models, possibly using different simulation codes.

\section*{Declarations}
\noindent {\bf Author contributions: \hspace{6pt}}
J.G.\ -- contributed to conceptualization, methodology, software development, investigation, visualization, writing, review and editing; S.W.\ -- contributed to methodology, software development, writing, review and editing; T.M.\ -- contributed to methodology, software development, investigation, visualization, writing, review and editing; P.M. \ -- contributed to software development, validation, writing, review and editing. M.P.\ -- contributed to conceptualization, methodology, software development, validation, writing, review and editing. 

\noindent {\bf Funding: \hspace{6pt}}
This research did not receive any specific grant from funding agencies in the public, commercial, or not-for-profit sectors.

\noindent {\bf Usage of AI in the present paper: \hspace{6pt}}
Being part of the research in this paper, Large Language Models (LLMs) have been used for code generation and as AI agents; apart from that, we used LLMs for spell-checking, text improvement and translation.
\vspace{6pt}\\

\renewcommand{\refname}{References\protect\footnote{For references with more than ten contributing authors, only the first ten authors are listed, followed by ``et al.'', to maintain readability and brevity in the list of references.}}
\bibliographystyle{elsarticle-num}
\bibliography{bibliographyDoc}

\newpage
\appendix
\setcounter{table}{0}

\section{Model List}
\label{sec:appendix:modelList}

\mytab{tab:modelList} lists all 84 mechanical models used in this study, organized by difficulty level.
\begin{table}[!h]
\centering
\caption{Complete list of mechanical models grouped by difficulty level.}
\label{tab:modelList}
\resizebox{0.95\linewidth}{!}{
\begin{tabular}{rl@{\hspace{1.5em}}rl}
\toprule
ID & model name & ID & model name \\
\midrule
\multicolumn{4}{c}{\cellcolor{green!15}\textbf{level 1 --- simple rigid MBS (mass points, springs, distance constraints)}} \\
1 & flying mass point & 16 & inverted single pendulum \\
2 & free fall mass point & 17 & double pendulum elastic spring \\
3 & single mass oscillator & 18 & n-pendulum elastic spring \\
4 & single mass oscillator with gravity & 19 & single pendulum / rigid body \\
5 & slider crank / point masses & 20 & mass point on rigid string \\
6 & pendulum with elastic string & 21 & mass point on elastic string \\
7 & mass oscillator with user function & 22 & link on two prismatic joints \\
8 & spinning disc & 23 & flying rigid body \\
9 & double mass oscillator & 24 & suspended rigid body \\
10 & n-mass oscillator & 25 & gyroscope on spherical joint \\
11 & single pendulum & 26 & prismatic joint system \\
12 & double pendulum & 27 & two mass points with springs \\
13 & n-pendulum & 28 & two mass points with distances \\
14 & spring coupled flying rigid bodies & 29 & rigid rotor simply supported \\
15 & torsional oscillator & 30 & double pendulum / rigid bodies \\
\midrule
\multicolumn{4}{c}{\cellcolor{blue!12}\textbf{level 2 --- complex rigid MBS (joints, contact, vehicles, robots)}} \\
31 & four-bar mechanism / point masses & 41 & sphere-sphere and sphere-ground impact \\
32 & disc rolling on ground & 42 & pendulum wall impacting \\
33 & elastic chain & 43 & Newton's cradle \\
34 & rigid rotor unbalanced & 44 & billiard with 4 spheres \\
35 & slider crank / rigid bodies & 45 & grid mesh and large sphere impact \\
36 & four-bar mechanism / rigid bodies & 46 & non-steerable scooter with ideal rolling \\
37 & parallel chord truss frame with springs & 47 & non-steerable four-wheeled car \\
38 & Pratt truss bridge frame with springs & 48 & gyroscope nutation \\
39 & sphere-sphere impact & 49 & 2R SCARA robot \\
40 & falling mass point with ground contact & 50 & RRR articulated robot with 3D cubic trajectory \\
\midrule
\multicolumn{4}{c}{\cellcolor{orange!20}\textbf{level 3a --- flexible body geometry (3D CAD / FEM parts)}} \\
51 & bearing block part & 62 & grinding bench spindle with disks part \\
52 & unbalanced rotor part & 63 & hinge leaf gap side part \\
53 & balanced rotor part & 64 & hinge leaf pin side part \\
54 & flexible beam part & 65 & mounting bracket part \\
55 & flexible beam fillets part & 66 & bearing block with pedestal part \\
56 & flexible beam hollow part & 67 & crank part simple \\
57 & flexible beam drilled part & 68 & rod part simple \\
58 & flexible table part & 69 & slider part simple \\
59 & wind turbine tower part & 70 & crank part advanced \\
60 & wind turbine rotor part & 71 & rod part advanced \\
61 & grinding bench housing part & 72 & slider part advanced \\
\midrule
\multicolumn{4}{c}{\cellcolor{violet!15}\textbf{level 3b --- flexible MBS assembly}} \\
73 & rotor in bearing block & 79 & simplified wind turbine \\
74 & rotor in bearing block with pedestal & 80 & grinding bench \\
75 & rotor in two bearing blocks with pedestal & 81 & door hinge \\
76 & flexible pendulum & 82 & mounting bracket free falling \\
77 & cantilever beam & 83 & flexible slider crank simple \\
78 & flexible table with load & 84 & flexible slider crank advanced \\
\bottomrule
\end{tabular}
}
\end{table}

\section{Complete LLM Evaluation Results}
\label{sec:appendix:llmResults}
\setcounter{table}{0} 

\mytab{tab:modelComparisonAll} lists the evaluation results for all evaluated LLMs on the combined set of 50 rigid-body task models.
\begin{table}[!h]
\centering
\caption{Evaluation results for all evaluated LLMs on the combined set of 50 rigid-body task models. Rates are given in \%. comp.\ = choose simulation components; solver = solver success; topol.\ = topology match; numsol = numerical solution match; graph = graph content match; overall = overall success. The horizontal rule separates proprietary models from open-weight models. $\dagger$gpt-oss LLMs using minimum reasoning effort (setting: ``low'').}
\label{tab:modelComparisonAll}
\resizebox{0.8\linewidth}{!}{
\begin{tabular}{l@{\hspace{-5pt}}rrrrrr}
\toprule
 & \multicolumn{6}{c}{\textbf{success rate (\%)}} \\
\cmidrule(lr){2-7}
\textbf{LLM} & \textbf{comp.} & \textbf{solver} & \textbf{topol.} & \textbf{numsol} & \textbf{graph} & \textbf{overall} \\
\midrule
Claude-Opus-4.8 & 99.4 & 99.7 & 99.2 & 96.1 & 93.2 & 91.4 \\
GPT-5.2 & 98.6 & 98.9 & 98.2 & 91.5 & 90.3 & 86.0 \\
\midrule
qwen3.6:27b & 95.8 & 94.0 & 91.0 & 83.3 & 83.3 & 82.1 \\
gemma4:31b & 97.0 & 95.9 & 94.3 & 84.3 & 84.0 & 81.1 \\
qwen3.5:27b & 96.6 & 96.0 & 93.9 & 83.1 & 83.5 & 81.1 \\
gpt-oss:120b\textsuperscript{$\dagger$} & 97.6 & 89.5 & 86.9 & 77.9 & 75.2 & 73.3 \\
qwen3.5:35b & 95.5 & 84.6 & 85.3 & 73.2 & 71.7 & 70.3 \\
gpt-oss:20b\textsuperscript{$\dagger$} & 97.8 & 87.8 & 85.3 & 72.0 & 70.0 & 67.6 \\
gemma4:26b & 99.9 & 84.7 & 84.5 & 70.2 & 66.6 & 66.0 \\
qwen3.5:122b & 86.6 & 88.3 & 87.6 & 76.4 & 74.1 & 62.5 \\
qwen3.6:35b & 92.7 & 79.3 & 77.0 & 69.0 & 65.8 & 60.7 \\
qwen2.5-coder:32b & 95.6 & 80.4 & 77.2 & 61.1 & 57.8 & 57.3 \\
ministral-3:14b & 99.2 & 88.2 & 81.4 & 60.8 & 54.5 & 54.4 \\
llama3.3:70b & 97.2 & 80.7 & 73.4 & 52.3 & 48.3 & 47.7 \\
llama3.1:70b & 88.6 & 75.9 & 71.9 & 54.1 & 49.5 & 47.3 \\
qwen2:72b & 89.9 & 80.2 & 72.7 & 44.9 & 40.0 & 38.7 \\
qwen3-coder:30b & 66.9 & 77.4 & 75.8 & 53.6 & 51.9 & 36.8 \\
phi4:latest & 75.2 & 73.2 & 67.7 & 47.9 & 46.7 & 36.7 \\
qwen3.5:4b & 78.5 & 76.3 & 62.0 & 43.4 & 39.3 & 33.8 \\
qwen3.5:9b & 78.1 & 77.3 & 69.6 & 54.6 & 47.4 & 33.2 \\
deepseek-r1:32b & 66.3 & 68.8 & 60.0 & 42.8 & 37.6 & 31.4 \\
glm-4.7-flash:q4\_K\_M & 85.8 & 72.5 & 49.8 & 35.6 & 29.2 & 29.0 \\
deepseek-r1:70b & 52.5 & 69.1 & 63.6 & 51.7 & 48.6 & 28.6 \\
nemotron-cascade-2:30b & 54.5 & 50.3 & 27.0 & 24.0 & 19.7 & 16.0 \\
deepseek-coder-v2:16b & 58.9 & 65.2 & 36.3 & 18.9 & 14.7 & 12.3 \\
codegemma:7b & 67.2 & 65.7 & 29.6 & 13.8 & 10.6 & 10.2 \\
dolphincoder:15b & 29.4 & 66.4 & 37.9 & 24.1 & 16.4 & 10.0 \\
nemotron3:33b & 30.9 & 36.0 & 15.0 & 10.5 & 9.3 & 9.0 \\
llama3.1:8b & 47.2 & 52.1 & 31.0 & 9.5 & 7.0 & 3.7 \\
codeqwen:7b & 41.3 & 50.5 & 14.5 & 4.5 & 2.7 & 2.7 \\
qwen2:7b & 61.5 & 47.2 & 23.6 & 6.2 & 2.8 & 2.0 \\
phi3:14b & 39.6 & 29.8 & 15.1 & 4.9 & 2.0 & 1.6 \\
qwen:72b & 35.7 & 1.4 & 1.4 & 1.4 & 1.4 & 1.4 \\
qwen:32b & 0.0 & 0.0 & 0.0 & 0.0 & 0.0 & 0.0 \\
\bottomrule
\end{tabular}
}
\end{table}

\end{document}